\documentclass{article} 
\usepackage{iclr2027_conference,times}
\setcitestyle{numbers,square,comma}

\usepackage{amsmath,amsfonts,bm}

\def\eqref#1{equation~\ref{#1}}

\def\1{\bm{1}}

\DeclareMathAlphabet{\mathsfit}{\encodingdefault}{\sfdefault}{m}{sl}
\SetMathAlphabet{\mathsfit}{bold}{\encodingdefault}{\sfdefault}{bx}{n}

\usepackage{hyperref}
\usepackage{url}
\usepackage{booktabs}   
\usepackage{multirow}   
\usepackage{graphicx}   
\usepackage{wrapfig}
\usepackage[table,dvipsnames]{xcolor}
\usepackage{diagbox}
\usepackage{tabularx}
\usepackage{amssymb}
\usepackage{amsmath}
\usepackage{subcaption}
\usepackage{pifont}
\usepackage{authblk}
\title{NeuroDETR: Multiscale Discrete Spatiotemporal Tokenization for Asynchronous EEG Event Detection}

\author[1,2,3]{Yaozhi Wen}
\author[1,4]{Kun Zhao}
\author[1]{Wei Wei}
\author[3,5]{Haoyang Chen}
\author[1]{Kangning Wang}
\author[1,3,4]{Yi Sun}
\author[1,2,3]{Shaowei Gu}
\author[3,6]{Dingkun Liu}
\author[1,*]{Shuang Qiu}
\author[1,2,3,4,*]{Huiguang He}

\affil[1]{Laboratory of Brain Atlas and Brain-Inspired Intelligence,
    State Key Laboratory of Brain Cognition and Brain-inspired Intelligence Technology,
    Institute of Automation, Chinese Academy of Sciences}

\affil[2]{School of Future Technology, University of Chinese Academy of Sciences}

\affil[3]{Zhongguancun Academy}

\affil[4]{School of Artificial Intelligence, University of Chinese Academy of Sciences}

\affil[5]{Wuhan University}

\affil[6]{Hubei Key Laboratory of Brain-inspired Intelligent Systems, School of Artificial Intelligence and Automation, Huazhong University of Science and Technology}

\affil[*]{Corresponding authors}
\affil[ ]{\texttt{ \{wenyaozhi2024, shuang.qiu, huiguang.he\}@ia.ac.cn}}
\iclrfinalcopy 
\begin{document}
\raggedbottom
\maketitle
\fancyhead{}
\renewcommand{\headrulewidth}{0pt}

\begin{abstract}
Asynchronous EEG event detection supports practical brain-computer interfaces by identifying event classes and temporal boundaries directly from continuous recordings without predefined event boundaries at inference time. The primary challenge lies in extracting event-related patterns embedded in ongoing background activity. To address this, we propose \textbf{NeuroDETR}, a discrete spatiotemporal multiscale framework that combines structured tokenization with event-aware feature refinement. Multiscale temporal codebooks capture brief responses and sustained dynamics, while Spatial-Pyramid VQ hierarchically encodes spatial structure across EEG channels. These complementary representations form a compact event memory for jointly classifying events and estimating their onsets and offsets. Before tokenization, we apply FOCAL, an event-aware sparse attention module that uses predicted event presence to guide temporal feature refinement and spatial aggregation, emphasizing likely event periods while preserving context across the recording. Experiments on eight EEG datasets that span diverse paradigms demonstrate state-of-the-art performance. These findings provide a basis for the development of BCI systems that recognize meaningful neural activity during continuous interaction and monitoring.
\end{abstract}

\section{Introduction}
Brain–computer interfaces (BCIs) establish direct communication pathways between neural activity and external devices by translating neural signals into actionable information for device control, human–computer interaction, and clinical decision-making~\citep{tang2023flexible,robinson2025application}. Achieving this goal depends on reliably identifying behaviorally or physiologically meaningful neural events in continuous electroencephalographic (EEG) recordings. Asynchronous EEG decoding addresses this challenge by jointly identifying event categories and localizing their onset and offset within the ongoing signal.

\begin{figure}[t]
  \centering
  \begin{minipage}{0.85\textwidth}
    \centering
    \begin{subfigure}[b]{0.57\linewidth}
      \centering
      \includegraphics[width=\linewidth]{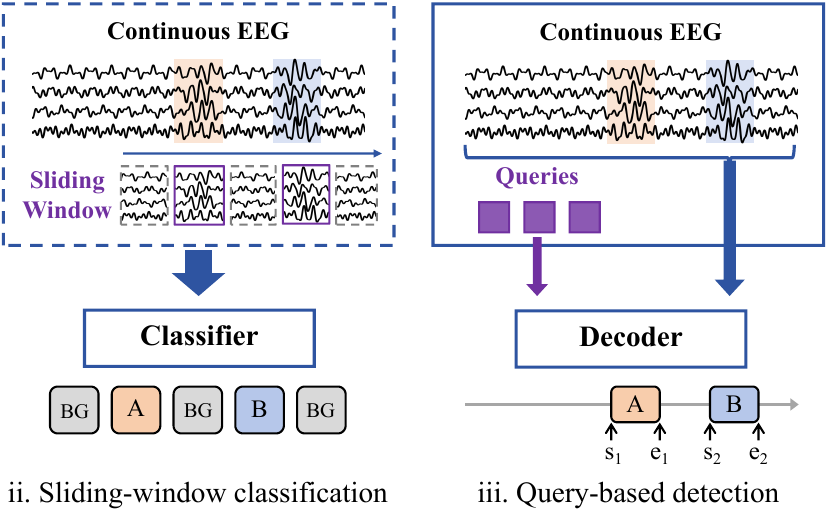}
      \caption{EEG decoding paradigms}
      \label{fig:analysis_paradigms}
    \end{subfigure}
    \hfill
    \begin{subfigure}[b]{0.38\linewidth}
      \centering
      \includegraphics[width=\linewidth]{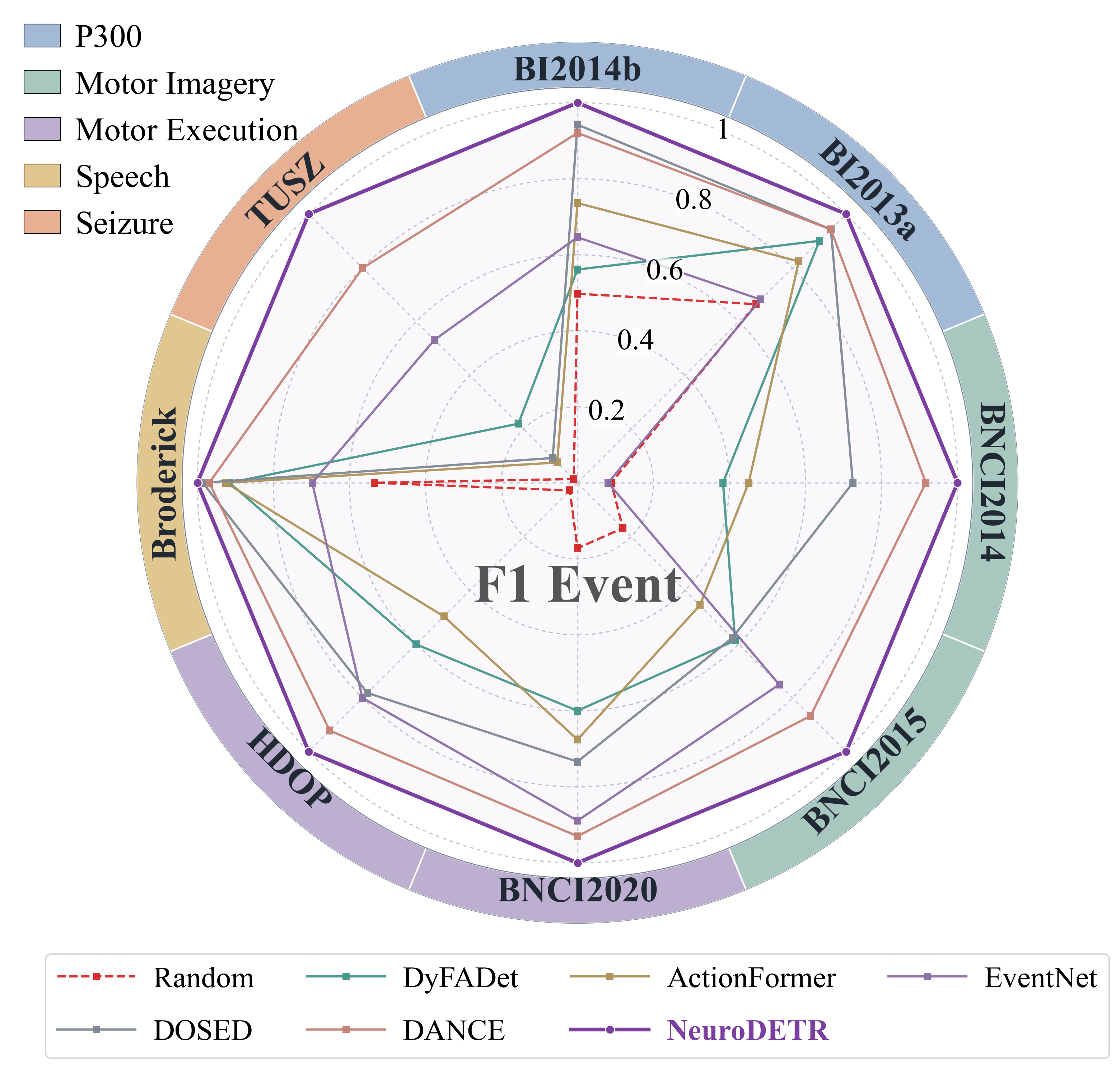}
      \caption{Relative performance}
      \label{fig:radar}
    \end{subfigure}
  \end{minipage}

  \caption{
    EEG decoding paradigms and relative performance.
    (a) Sliding-window classification and query-based event detection.
    (b) Performance of NeuroDETR and six baselines across
    eight datasets. Scores on each dataset are normalized
    by NeuroDETR's score (NeuroDETR = 1).
  }
\end{figure}

As shown in Figure~\ref{fig:analysis_paradigms} (i), a straightforward approach to asynchronous decoding applies sliding-window classifiers to continuous EEG, assigning each window an event or background label~\citep{wu2024motor,lopes2019online}. However, the mismatch between fixed-length windows and variable event durations and boundaries can fragment detections or merge adjacent events, making performance sensitive to window settings and post-processing rules~\citep{seeuws2024avoiding}. As illustrated in Figure~\ref{fig:analysis_paradigms} (ii), query-based detection addresses this mismatch by directly mapping continuous EEG to an unordered set of events, each characterized by its category, onset, and offset. 

The effectiveness of this approach depends on how well the EEG features supplied to the decoder preserve event-specific information. Existing methods use cross-attention to aggregate information from long feature sequences into a fixed set of continuous latent tokens, but do not explicitly organize event cues across temporal scales and spatial regions~\citep{jaegle2021perceiver,levy2026dance}. Across EEG datasets and paradigms, events vary in duration, occurrence frequency, and spatial distribution. Their responses are also embedded in ongoing background activity, requiring feature aggregation to preserve weak or transient event cues.

In this study, we propose NeuroDETR, a multiscale discrete spatiotemporal framework for asynchronous EEG event detection. NeuroDETR uses vector quantization to organize continuous features into reusable learned codewords. To accommodate differences across datasets and paradigms, multiscale temporal VQ represents both brief responses and sustained dynamics, while Spatial-Pyramid VQ encodes channel information at complementary spatial levels. These representations form a compact event memory that supports joint localization and classification within a unified detection architecture.

To guide this representation toward event-related activity, we introduce FOCAL, an event-aware sparse attention module. FOCAL uses predicted event presence to guide temporal feature refinement before quantization and temporal pooling within the spatial branch, while retaining cross-window context. This shared event guidance connects the temporal and spatial representations, supplying the DETR decoder with event-aware multiscale information for predicting event classes and boundaries.
We evaluate NeuroDETR on eight EEG datasets spanning P300, motor imagery, motor execution, speech, and seizure detection. NeuroDETR achieves the highest event-level F1 among the evaluated methods on all eight datasets, demonstrating its effectiveness across EEG paradigms with diverse event characteristics. By improving neural event detection in continuous EEG, NeuroDETR supports the development of practical BCI applications.

Our main contributions are summarized as follows:

\begin{itemize}
    \item We propose \textbf{NeuroDETR}, which jointly predicts event classes, onsets, and offsets from continuous EEG without predefined event boundaries. At its core, vector quantization organizes EEG features into a compact event memory, with multiscale temporal and hierarchical spatial codebooks accommodating diverse event characteristics across paradigms.

   \item We introduce \textbf{FOCAL}, an event-aware sparse attention module that concentrates local feature refinement on likely event periods while preserving cross-window context. By sharing event-presence estimates with spatial aggregation, FOCAL directs both temporal and spatial tokenization toward event-relevant information.

    \item Experiments on eight EEG datasets spanning diverse paradigms demonstrate state-of-the-art performance. Ablation studies support the complementary contributions of spatiotemporal tokenization and event-aware refinement.
\end{itemize}

\section{Related Work}

\subsection{Asynchronous EEG Decoding}
Asynchronous EEG decoding identifies events in continuous signals without known onset times.
Early systems estimate control states or classify sliding windows, using probabilistic aggregation or temporal decision rules~\citep{zhang2008asynchronous,lopes2019online,wu2024motor}.
Recovering event instances typically requires temporal post-processing.
DOSED and EventNet directly localize events through single-stage detection and center--duration regression, respectively~\citep{chambon2019dosed,seeuws2024avoiding}.
DETRtime uses event queries and bipartite matching for set prediction~\citep{wolf2022deep}, while a frozen foundation model has also been adapted for dense EEG segmentation~\citep{ma2026adapting}.
DANCE combines Perceiver-style resampling~\citep{jaegle2021perceiver} with query decoding to predict event classes and temporal boundaries~\citep{levy2026dance}.

\subsection{Discrete Spatiotemporal EEG Representations}
Discrete EEG tokens support EEG-to-text translation in DeWave~\citep{duan2023dewave} and pretraining or multimodal decoding in LaBraM, VQ-MTM, and NeuroLM~\citep{jiang2024large,gui2024vector,jiang2025neurolm}.
Recent tokenizers learn single-channel time--frequency motifs and decoupled temporal--spectral vocabularies~\citep{pradeepkumar2026tokenizing,ma2025codebrain}.
Related approaches model EEG through electrode graphs, multiscale temporal filters, and regional and global spatial representations~\citep{ding2022tsception,gao2024graph}, or use sparse attention for motor imagery decoding~\citep{liu2024empt,miao2025satrans}.
THD-BAR constructs hierarchical spatial tokens for autoregressive learning~\citep{yang2026thd}, while NeuroRVQ combines multiscale temporal convolutions with residual vector quantization for biosignal reconstruction and foundation-model training~\citep{barmpas2025neurorvq}.

NeuroDETR uses shared event-presence estimates to route local temporal attention and weight temporal pooling of regional features before multiscale temporal and hierarchical spatial quantization.
The resulting memory supports event-set prediction.

\section{Method}
\label{sec:method}

\subsection{Problem Formulation and Overview}

We study asynchronous event detection from continuous EEG. Given $\mathbf X\in\mathbb R^{C\times T}$ with $C$ channels and $T=f_sW$ samples, where $f_s$ is the sampling rate and $W$ the window duration, the target is an unordered event set:
\begin{equation}
    \mathcal Y=\{(c_i,s_i,e_i)\}_{i=1}^{M},
    \quad c_i\in\{1,\ldots,K\},\quad 0\leq s_i<e_i\leq1.
    \label{eq:method_targets}
\end{equation}
Here, $M$ varies across windows, and boundaries are normalized by $W$. NeuroDETR adopts a DETR-style detection framework to predict events in parallel using $N_e$ learnable event queries, with class $0$ denoting no event.

\begin{figure*}[!t]
    \centering
    \includegraphics[width=\textwidth]{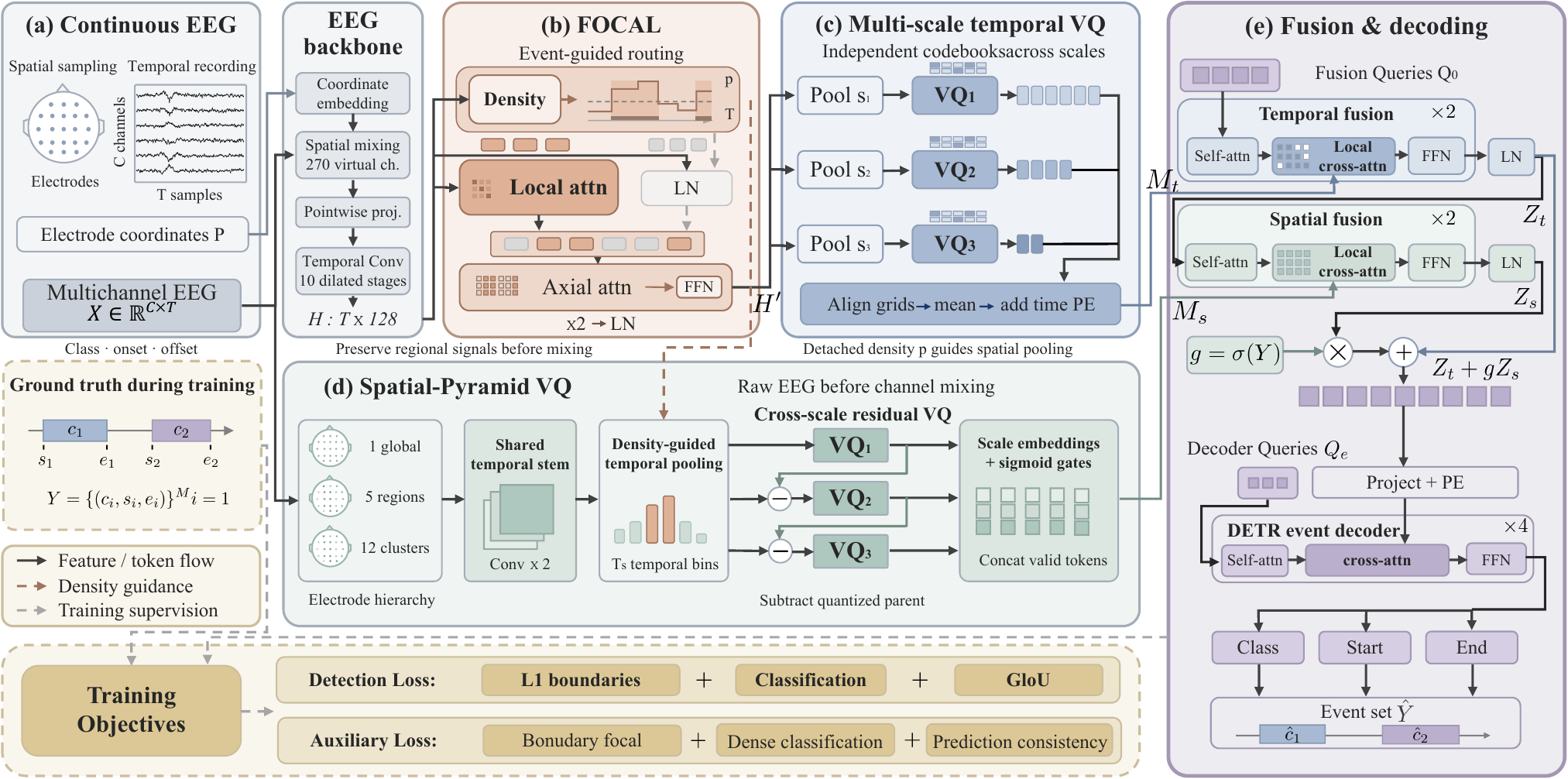}
\caption{
Overview of NeuroDETR.
\textbf{(a)} Continuous EEG input.
\textbf{(b)} FOCAL uses event-presence estimates to guide local attention.
\textbf{(c)} Temporal features are quantized at multiple resolutions.
\textbf{(d)} Spatial-Pyramid VQ quantizes EEG features
at global, regional, and local spatial levels.
\textbf{(e)} A query decoder predicts event classes and boundaries
from fused spatiotemporal memory.
Detection and auxiliary losses provide training supervision.
}
    \label{fig:framework}
\end{figure*}

As shown in Figure~\ref{fig:framework}, a convolutional backbone extracts dense features, which FOCAL refines through event-guided local attention and cross-window interaction. Temporal VQ encodes dynamics at multiple resolutions, while spatial VQ retains regional electrode information before channel mixing. Query-based fusion combines both memories for event localization and classification~\citep{carion2020end}.

\subsection{Dense EEG Encoding}

We adapt SimpleConv from D{\'e}fossez et al.~\citep{defossez2023decoding}. Spatial attention uses Fourier embeddings of normalized electrode coordinates $\mathbf P\in\mathbb R^{C\times2}$ to mix the inputs into 270 virtual channels. A pointwise projection and ten dilated temporal convolutional stages yield
\begin{equation}
    \mathbf H=f_{\mathrm{enc}}(\mathbf X,\mathbf P)
    \in\mathbb R^{T\times D},\qquad D=128.
    \label{eq:dense_encoding}
\end{equation}
The backbone preserves temporal resolution, supplying features for FOCAL and auxiliary boundary supervision. Architectural details appear in Appendix~\ref{app:backbone}.

\subsection{FOCAL: Event-Aware Sparse Attention}

FOCAL combines event-guided local attention with axial interaction across windows.

\paragraph{Event-guided routing.}
We partition $\mathbf H$ into $N_w=\lceil T/w\rceil$ windows of $w$ tokens, padding the last window with zeros. A density head $g_\phi$, comprising layer normalization, two linear layers, and GELU, predicts event presence from mean-pooled features:
\begin{equation}
    \ell_n=g_\phi\!\left(\frac{1}{w}\sum_{t\in\mathcal W_n}\mathbf h_t\right),
    \qquad p_n=\sigma(\ell_n).
    \label{eq:method_density}
\end{equation}
Window targets indicate overlap with an annotated event:
\begin{equation}
    y_n=\mathbb 1\!\left[
        \exists i:\ e_i>\frac{n}{N_w},\
        s_i<\frac{n+1}{N_w}
    \right],
    \quad n=0,\ldots,N_w-1.
    \label{eq:method_density_target}
\end{equation}
We train $g_\phi$ with binary cross-entropy:
\begin{equation}
    \mathcal L_{\mathrm{den}}
    =-\frac{1}{N_w}\sum_{n=0}^{N_w-1}
    \left[y_n\log p_n+(1-y_n)\log(1-p_n)\right].
    \label{eq:method_density_loss}
\end{equation}
Windows are selected for local attention by
\begin{equation}
\mathcal I_{\mathrm{evt}} = \{n:p_n>\tau\} \cup \operatorname{TopK}(\boldsymbol\ell,k_{\min}), \quad
m_n = \mathbb 1[n\in\mathcal I_{\mathrm{evt}}].
\label{eq:method_routing}
\end{equation}
The default uses $\tau=0.4$ and $k_{\min}=1$. The density map and routing mask are computed once and shared across both FOCAL blocks. Hard routing leaves $g_\phi$ supervised by $\mathcal L_{\mathrm{den}}$.

\paragraph{Local--axial interaction.}
Selected windows receive local self-attention, and the others receive background normalization:
\begin{equation}
    \widetilde{\mathbf H}_n=
    \begin{cases}
        \mathbf H_n+\operatorname{MHSA}(\operatorname{LN}(\mathbf H_n)),
        &m_n=1,\\
        \operatorname{LN}_{\mathrm{bg}}(\mathbf H_n),&m_n=0.
    \end{cases}
    \label{eq:method_local}
\end{equation}
For $\mathbf H_n\in\mathbb R^{w\times D}$, axial attention~\citep{ho2019axial} connects corresponding token positions across all windows by reshaping
\begin{equation}
    [B,N_w,w,D]\longrightarrow[Bw,N_w,D],
    \label{eq:method_axial}
\end{equation}
where $B$ is the batch size. Each block applies axial attention and a feed-forward network with pre-normalization and residual connections. Two blocks, final normalization, and padding removal yield $\mathbf H'\in\mathbb R^{T\times D}$ for temporal VQ.

\subsection{Spatiotemporal VQ Tokenization}
\label{subsec:spatial_vq}

Temporal and spatial branches use independent, scale-specific codebooks to encode temporal dynamics and channel-group structure.

\paragraph{Quantization and codebook learning.}
Following VQ representation learning~\citep{van2017neural}, we assign $\mathbf z$ to its nearest codeword in $\mathcal E=\{\mathbf e_j\}_{j=1}^{V}$:
\begin{equation}
    k^\star=\arg\min_j\|\mathbf z-\mathbf e_j\|_2^2,
    \qquad
    Q(\mathbf z)=\mathbf z+
    \operatorname{sg}[\mathbf e_{k^\star}-\mathbf z].
    \label{eq:method_quantization}
\end{equation}
$\operatorname{sg}[\cdot]$ denotes stop-gradient, so $Q$ returns
the codeword and passes gradients to $\mathbf z$.
Each quantizer incurs a commitment penalty
$\beta\operatorname{MSE}(\mathbf z,\operatorname{sg}[\mathbf e_{k^\star}])$.
Codebooks are initialized from features and updated by exponential
moving averages (EMA), with underused codes reinitialized.
Commitment losses are averaged across scales within each branch,
using only valid tokens for spatial VQ.

\subsubsection{MultiScale Temporal VQ}

At each scale $s\in\mathcal S_{\mathrm t}$, average pooling with kernel and stride $s$ retains $\lceil T/s\rceil$ tokens, including the final partial bin:
\begin{equation}
    \mathbf H_s=\operatorname{AvgPool}_s(\mathbf H'),
    \qquad
    \mathbf q_s^{\mathrm t}=Q^{\mathrm t,s}(\mathbf H_s).
    \label{eq:method_temporal_vq}
\end{equation}
We linearly interpolate the codes to the finest pooled grid, $L_{\mathrm m}=\max_s\lceil T/s\rceil$, and average them before adding sinusoidal positional embeddings:
\begin{equation}
    \mathbf M^{\mathrm t}
    =\frac{1}{|\mathcal S_{\mathrm t}|}
    \sum_{s\in\mathcal S_{\mathrm t}}
    \operatorname{Up}_{L_{\mathrm m}}(\mathbf q_s^{\mathrm t})
    +\operatorname{PE}_{L_{\mathrm m}}.
    \label{eq:method_temporal_memory}
\end{equation}
A two-layer MLP $r_\psi$ reconstructs dense features from the averaged codes resampled to length $T$:
\begin{equation}
    \mathcal L_{\mathrm{rec}}
    =\operatorname{MSE}\!\left(
        r_\psi\!\left[
            \frac{1}{|\mathcal S_{\mathrm t}|}
            \sum_{s\in\mathcal S_{\mathrm t}}
            \operatorname{Up}_T(\mathbf q_s^{\mathrm t})
        \right],
        \operatorname{sg}[\mathbf H']
    \right).
    \label{eq:method_reconstruction}
\end{equation}
Reconstruction uses codes before positional embedding and memory fusion.

\subsubsection{Spatial-Pyramid VQ}

Spatial-Pyramid VQ aggregates raw EEG channels before backbone mixing at three levels: one whole-head group, five coarse groups (frontal, temporal, central, parietal, and occipital), and twelve fine electrode clusters:
\begin{equation}
    \mathbf R_\ell=\mathbf A_\ell\mathbf X
    \in\mathbb R^{N_\ell\times T},\qquad
    (N_1,N_2,N_3)=(1,5,12).
    \label{eq:method_regions}
\end{equation}
Each nonempty row of $\mathbf A_\ell$ averages available channels assigned by channel name. A shared stem with two strided temporal convolutions and GELU produces $\mathbf F_{\ell,r,u}\in\mathbb R^D$, indexed by region $r$ and stem time step $u$. Details of the hierarchical electrode grouping are provided
in Appendix~\ref{app:brain_region_grouping}.

\paragraph{Density-guided pooling.}
FOCAL probabilities $\mathbf d=\operatorname{sg}[\sigma(\boldsymbol\ell)]$ are linearly interpolated to the stem grid as $\widetilde d_u$ and weight each group's features within $T_{\mathrm s}$ temporal bins:
\begin{equation}
    \mathbf E_{\ell,r,b}=
    \frac{
        \sum_u\kappa(b,u)\widetilde d_u\mathbf F_{\ell,r,u}
    }{
        \max\!\left(
            \sum_u\kappa(b,u)\widetilde d_u,\epsilon
        \right)
    }.
    \label{eq:method_spatial_pooling}
\end{equation}
Here, $b$ indexes output bins, $\kappa$ is a row-normalized triangular kernel centered on each bin, and $\epsilon=10^{-6}$. The detached weights emphasize likely event periods without updating the density estimator.

\paragraph{Cross-scale residual quantization.}
The whole-head descriptor is quantized directly, while each finer level quantizes its residual relative to its parent's code:
\begin{equation}
    \mathbf u_\ell=
    \begin{cases}
        \mathbf E_1,&\ell=1,\\
        \mathbf E_\ell-
        \operatorname{Broadcast}_\ell(
            \mathbf q_{\ell-1}^{\mathrm s}
        ),&\ell>1,
    \end{cases}
    \qquad
    \mathbf q_\ell^{\mathrm s}
    =Q^{\mathrm s,\ell}(\mathbf u_\ell).
    \label{eq:method_spatial_residual}
\end{equation}
$\operatorname{Broadcast}_\ell$ assigns each child the immediately preceding level's quantized parent output at the same temporal bin. Learnable scale embeddings $\mathbf a_\ell$ and sigmoid gates $g_\ell=\sigma(\eta_\ell)$ form the spatial memory:
\begin{equation}
    \mathbf M^{\mathrm s}
    =\mathop{\Vert}_{\ell=1}^{3}
    \operatorname{Flatten}_{\mathrm{valid}}\!\left[
        g_\ell(\mathbf q_\ell^{\mathrm s}+\mathbf a_\ell)
    \right].
    \label{eq:method_spatial_memory}
\end{equation}
$\operatorname{Flatten}_{\mathrm{valid}}$ retains all temporal bins of nonempty channel groups.

\subsection{Event Memory Fusion and Set Prediction}
\label{subsec:fusion}

\paragraph{Temporal fusion and gated spatial integration.}
Starting from $N_q$ learnable queries initialized with sinusoidal embeddings, temporal fusion alternates global self-attention, sliding-window cross-attention (SWA), and feed-forward refinement:
\begin{equation}
\mathbf Q \leftarrow \mathbf Q + \operatorname{MHSA}(\operatorname{LN}(\mathbf Q)), \quad
\mathbf Q \leftarrow \mathbf Q + \operatorname{SWA}(\operatorname{LN}(\mathbf Q), \mathbf M^{\mathrm t}), \quad
\mathbf Q \leftarrow \mathbf Q + \operatorname{FFN}(\operatorname{LN}(\mathbf Q)).
\label{eq:method_fusion_layer}
\end{equation}
Cross-attention spans are centered proportionally along the memory. The normalized output $\mathbf Z_{\mathrm t}=\operatorname{LN}[\mathcal F_{\mathrm t}(\mathbf Q_0,\mathbf M^{\mathrm t})]$ initializes an independently parameterized spatial Transformer with the same structure, whose output enters through a gated residual:
\begin{equation}
    \mathbf Z=\mathbf Z_{\mathrm t}
    +g\operatorname{LN}[\mathcal F_{\mathrm s}(\mathbf Z_{\mathrm t},\mathbf M^{\mathrm s})],
    \qquad g=\sigma(\gamma).
    \label{eq:method_gated_fusion}
\end{equation}
Each stream has two layers and four heads, yielding memory $\mathbf Z\in\mathbb R^{N_q\times D}$. The spatial gate is initialized at $g=0.02$.

\paragraph{Set-based event decoding.}
A four-layer, four-head DETR decoder~\citep{carion2020end} applies $N_e$ event queries to the fused memory, projected from 128 to 256 dimensions and augmented with sinusoidal embeddings. Query features $\mathbf o_j$ predict classes and boundaries:
\begin{equation}
    \widehat{\mathbf p}_j=\operatorname{softmax}(\operatorname{Cls}(\mathbf o_j)),\quad
    \hat s_j=\sigma(\operatorname{Start}(\mathbf o_j)),\quad
    \hat e_j=\sigma(\operatorname{End}(\mathbf o_j)).
    \label{eq:method_event_heads}
\end{equation}
We discard queries whose highest-probability class is background and multiply the remaining boundaries by $W$. Here, $N_q$ controls memory length and $N_e$ the candidate-event budget.

\subsection{Training Objectives}

\paragraph{Detection Loss.}
Hungarian matching~\citep{kuhn1955hungarian} assigns ground-truth events to predicted queries using a matching cost based on class probabilities and temporal IoU.
The classification loss $\mathcal L_{\mathrm{cls}}$ applies cross-entropy supervision to all queries, assigning event labels to matched queries and the background label to unmatched queries.
For matched queries, endpoint regression loss $\mathcal L_{\ell_1}$ penalizes absolute onset and offset errors.
Temporal generalized IoU loss $\mathcal L_{\mathrm{GIoU}}$~\citep{rezatofighi2019generalized} complements endpoint regression by supervising interval overlap and relative position.
The detection objective is
\begin{equation}
    \mathcal L_{\mathrm{det}}
    = \lambda_{\ell_1}\mathcal L_{\ell_1}
    + \lambda_{\mathrm{cls}}\mathcal L_{\mathrm{cls}}
    + \lambda_{\mathrm{giou}}\mathcal L_{\mathrm{GIoU}}.
    \label{eq:method_detection_loss}
\end{equation}

\paragraph{Auxiliary Loss.}
We complement set prediction with full-resolution boundary supervision,
dense classification, and prediction consistency.
A boundary head applied to the backbone features $\mathbf H$ predicts
class-agnostic onset and offset heatmaps.
The targets are pointwise maxima of Gaussians centered at annotated
boundaries.
Sigmoid focal loss $\mathcal L_{\mathrm{bnd}}$ supervises these predictions
while downweighting easy non-boundary positions.

For dense classification and prediction consistency, we follow the
auxiliary-supervision strategy of DANCE~\citep{levy2026dance}.
Dense cross-entropy loss $\mathcal L_{\mathrm{dense}}$ supervises class
predictions at each position of the fused memory $\mathbf Z$.
To construct the set-derived predictions, we accumulate query class
probabilities over their discretized predicted intervals and normalize
the accumulated scores at each covered grid position.
Uncovered positions are assigned zero vectors.
We apply componentwise clamping to both predictions without
renormalization and compute a KL-style regularizer
$\mathcal L_{\mathrm{cons}}$.
At covered grid positions, this term encourages agreement
between dense and set-derived predictions.
At uncovered positions, the clamped zero vectors induce
an entropy regularization effect on the dense predictions.

The complete event-supervision objective is
\begin{equation}
\begin{aligned}
    \mathcal L_{\mathrm{event}}
    ={}&\boldsymbol{\mathcal L_{\mathrm{det}}}
    +\lambda_{\mathrm{bnd}}\mathcal L_{\mathrm{bnd}}
    +\lambda_{\mathrm{dense}}\mathcal L_{\mathrm{dense}}
    +\lambda_{\mathrm{cons}}\mathcal L_{\mathrm{cons}}.
\end{aligned}
    \label{eq:method_total_loss}
\end{equation}
More details on the training objectives are provided in the appendix.


\section{Experiments}
\subsection{Datasets}
We evaluate our method on eight EEG datasets spanning five task domains with diverse temporal characteristics. These datasets cover densely occurring and potentially overlapping P300 events (BI2013a~\citep{congedo2011brain} and BI2014b~\citep{korczowski_2019_3267302}), fine-grained high-frequency speech-related events (Broderick~\citep{broderick2018electrophysiological,broderick2019data}), sparse and long-duration motor imagery events (BNCI2014~\citep{tangermann2012review} and BNCI2015~\citep{faller2012autocalibration}), motor execution events (BNCI2020~\citep{schwarz2020analyzing} and HDOP~\citep{lomele2026high}), and rare epileptic seizures with durations ranging from several seconds to several minutes (TUSZ (v2.0.6)~\cite{shah2018temple}). This diversity enables a comprehensive evaluation of whether a unified set-prediction framework can generalize across variations in event frequency, duration, category complexity, and recording configurations. Further details on the datasets are provided in Appendix \ref{sec:dataset_implementation}.

\subsection{Baselines} 
We compare NeuroDETR with four dense temporal detection frameworks, EventNet~\citep{seeuws2024avoiding}, ActionFormer~\citep{zhang2022actionformer}, DyFADet~\citep{yang2024dyfadet}, and DOSED~\citep{chambon2019dosed}, and the query-based set-prediction framework DANCE~\citep{levy2026dance}. We also include a dataset-aware Random event generator as a lower bound. For each learned baseline, we adapt the input representation, classification head, and temporal decoding procedure to EEG event detection while retaining its original detection mechanism. All methods use the same data splits and event-level evaluation protocol. For all dense detectors, we apply a confidence threshold of 0.05 and non-maximum suppression (NMS) with a temporal IoU threshold of 0.5, retaining at most $N_e$ predictions per window. Further implementation details appear in Appendix~\ref{sec:baseline_implementation}.

\subsection{Evaluation Metrics}

We report $F1_{\mathrm{sample}}$ and $F1_{\mathrm{event}}$.
The former measures sample-wise classification quality through macro-averaged F1, while the latter jointly evaluates event detection, classification, and localization.
Under one-to-one matching, a predicted event counts as a true positive when
\begin{equation}
    \operatorname{IoU}(\hat I_j,I_i)>\tau_{\mathrm{IoU}}
    \quad\text{and}\quad
    \hat c_j=c_i,
    \label{eq:event_matching}
\end{equation}
where $\hat I_j$ and $I_i$ denote the predicted and ground-truth intervals, and $\hat c_j$ and $c_i$ their respective classes.
We set $\tau_{\mathrm{IoU}}=0.5$ for the main evaluation.
The two metrics are computed as
\begin{equation}
\begin{aligned}
    F1_{\mathrm{sample}}
    &=\frac{1}{|\mathcal C|}\sum_{c\in\mathcal C}
    \frac{2TP_c^{\mathrm{s}}}
    {2TP_c^{\mathrm{s}}+FP_c^{\mathrm{s}}+FN_c^{\mathrm{s}}},\\
    F1_{\mathrm{event}}
    &=\frac{2TP^{\mathrm{e}}}
    {2TP^{\mathrm{e}}+FP^{\mathrm{e}}+FN^{\mathrm{e}}}.
\end{aligned}
    \label{eq:evaluation_metrics}
\end{equation}
Here, $\mathcal C$ is the evaluated class set, and superscripts $\mathrm{s}$ and $\mathrm{e}$ denote counts over time samples and event instances, respectively.

\renewcommand{\arraystretch}{1.1}
\begin{table}[!t]
\centering
\caption{
Results on eight EEG datasets: mean $\pm$ SD over five
cross-validation folds, except TUSZ, which uses a single
result on its predefined split.
Best and second-best scores are \textbf{bold} and
\underline{underlined}, respectively.
}
\resizebox{\textwidth}{!}{%
\begin{tabular}{c|l|cccccccc}
\toprule
&
\multirow{2}{*}[-0.6ex]{\textbf{Method}}
& \multicolumn{2}{c}{\textbf{P300}}
& \multicolumn{2}{c}{\textbf{Motor Imagery}}
& \multicolumn{2}{c}{\textbf{Motor Execution}}
& \multicolumn{1}{c}{\textbf{Speech}}
& \multicolumn{1}{c}{\textbf{Seizure}} \\
\cmidrule(lr){3-4}
\cmidrule(lr){5-6}
\cmidrule(lr){7-8}
\cmidrule(lr){9-9}
\cmidrule(lr){10-10}
&
& BI2014b & BI2013a
& BNCI2014 & BNCI2015
& BNCI2020 & HDOP
& Broderick
& TUSZ \\
\midrule
\multirow{7}{*}{%
\rotatebox[origin=c]{90}{\textcolor{black}{\textbf{F1 Sample}}}}
& \cellcolor{gray!30} Random
& \cellcolor{gray!30} $35.46_{\pm 0.60}$
& \cellcolor{gray!30} $40.40_{\pm 0.36}$
& \cellcolor{gray!30} $20.06_{\pm 0.54}$
& \cellcolor{gray!30} $33.75_{\pm 1.12}$
& \cellcolor{gray!30} $33.50_{\pm 0.50}$
& \cellcolor{gray!30} $25.11_{\pm 0.39}$
& \cellcolor{gray!30} $48.94_{\pm 0.15}$
& \cellcolor{gray!30} $19.67$ \\

& DyFADet
& $34.84_{\pm 9.46}$ & $\underline{55.08}_{\pm 2.22}$
& $34.58_{\pm 20.67}$ & $63.68_{\pm 3.52}$
& $45.99_{\pm 22.05}$ & $51.40_{\pm 1.23}$
& $47.78_{\pm 2.66}$ & $25.60$ \\

& ActionFormer
& $52.18_{\pm 1.65}$ & $54.71_{\pm 2.29}$
& $42.38_{\pm 4.85}$ & $53.04_{\pm 1.29}$
& $53.50_{\pm 2.16}$ & $46.63_{\pm 1.80}$
& $42.75_{\pm 1.17}$ & $22.56$ \\

& EventNet
& $44.04_{\pm 2.74}$ & $51.61_{\pm 5.88}$
& $\underline{53.64}_{\pm 3.50}$ & $\underline{73.94}_{\pm 4.02}$
& $\textbf{59.91}_{\pm 2.94}$ & $57.52_{\pm 3.82}$
& $\underline{52.18}_{\pm 0.61}$ & $24.52$ \\

& DOSED
& $38.56_{\pm 2.30}$ & $44.75_{\pm 3.01}$
& $46.06_{\pm 2.98}$ & $60.85_{\pm 3.55}$
& $55.36_{\pm 2.93}$ & $\underline{64.15}_{\pm 1.69}$
& $43.30_{\pm 0.44}$ & $18.27$ \\

& DANCE
& $\underline{56.24}_{\pm 3.04}$ & $54.73_{\pm 2.41}$
& $53.46_{\pm 6.44}$ & $73.36_{\pm 2.54}$
& $55.79_{\pm 2.56}$ & $61.74_{\pm 3.37}$
& $50.31_{\pm 2.71}$ & $\underline{27.78}$ \\

\cline{2-10}
& \cellcolor{SkyBlue!30} \textbf{NeuroDETR}
& \cellcolor{SkyBlue!30} $\textbf{59.61}_{\pm 3.12}$
& \cellcolor{SkyBlue!30} $\textbf{57.48}_{\pm 4.02}$
& \cellcolor{SkyBlue!30} $\textbf{55.63}_{\pm 5.17}$
& \cellcolor{SkyBlue!30} $\textbf{75.82}_{\pm 3.30}$
& \cellcolor{SkyBlue!30} $\underline{59.05}_{\pm 2.80}$
& \cellcolor{SkyBlue!30} $\textbf{66.32}_{\pm 3.35}$
& \cellcolor{SkyBlue!30} $\textbf{53.80}_{\pm 0.68}$
& \cellcolor{SkyBlue!30} $\textbf{32.26}$ \\

\midrule
\multirow{7}{*}{%
\rotatebox[origin=c]{90}{\textcolor{black}{\textbf{F1 Event}}}}
& \cellcolor{gray!30} Random
& \cellcolor{gray!30} $22.58_{\pm 0.58}$
& \cellcolor{gray!30} $37.55_{\pm 0.50}$
& \cellcolor{gray!30} $4.05_{\pm 0.56}$
& \cellcolor{gray!30} $11.44_{\pm 0.76}$
& \cellcolor{gray!30} $7.21_{\pm 0.54}$
& \cellcolor{gray!30} $1.51_{\pm 0.55}$
& \cellcolor{gray!30} $20.69_{\pm 0.06}$
& \cellcolor{gray!30} $0.35$ \\

& DyFADet
& $25.38_{\pm 2.58}$ & $50.86_{\pm 1.86}$
& $17.40_{\pm 15.51}$ & $39.76_{\pm 3.95}$
& $25.10_{\pm 12.94}$ & $31.80_{\pm 1.14}$
& $35.49_{\pm 0.43}$ & $5.51$ \\

& ActionFormer
& $33.29_{\pm 1.72}$ & $46.53_{\pm 1.34}$
& $20.47_{\pm 6.14}$ & $30.88_{\pm 2.11}$
& $28.28_{\pm 1.92}$ & $26.30_{\pm 1.30}$
& $35.86_{\pm 0.28}$ & $1.89$ \\

& EventNet
& $29.22_{\pm 0.59}$ & $38.52_{\pm 1.44}$
& $30.67_{\pm 4.63}$ & $50.86_{\pm 5.24}$
& $37.21_{\pm 2.21}$ & $42.34_{\pm 4.42}$
& $27.05_{\pm 0.76}$ & $13.33$ \\

& DOSED
& $\underline{42.62}_{\pm 2.64}$ & $\underline{53.28}_{\pm 0.88}$
& $32.92_{\pm 3.75}$ & $39.08_{\pm 0.93}$
& $30.71_{\pm 2.98}$ & $41.37_{\pm 2.16}$
& $\underline{38.22}_{\pm 0.50}$ & $2.31$ \\

& DANCE
& $41.63_{\pm 2.12}$ & $53.26_{\pm 0.75}$
& $\underline{41.65}_{\pm 6.96}$ & $\underline{58.78}_{\pm 3.37}$
& $\underline{38.92}_{\pm 3.64}$ & $\underline{47.48}_{\pm 4.46}$
& $37.55_{\pm 0.83}$ & $\underline{20.03}$ \\

\cline{2-10}
& \cellcolor{SkyBlue!30} \textbf{NeuroDETR}
& \cellcolor{SkyBlue!30} $\textbf{45.23}_{\pm 1.22}$
& \cellcolor{SkyBlue!30} $\textbf{56.50}_{\pm 0.97}$
& \cellcolor{SkyBlue!30} $\textbf{45.55}_{\pm 4.96}$
& \cellcolor{SkyBlue!30} $\textbf{67.77}_{\pm 3.99}$
& \cellcolor{SkyBlue!30} $\textbf{41.85}_{\pm 3.83}$
& \cellcolor{SkyBlue!30} $\textbf{52.91}_{\pm 3.33}$
& \cellcolor{SkyBlue!30} $\textbf{38.76}_{\pm 0.57}$
& \cellcolor{SkyBlue!30} $\textbf{25.06}$ \\
\bottomrule
\end{tabular}%
}
\label{llava-main-result}
\end{table}


\renewcommand{\arraystretch}{1}
\begin{table}[t]
\centering
\caption{
Core component ablations: $F1_{\mathrm{event}}$ on eight EEG
datasets (mean $\pm$ SD over five cross-validation folds;
TUSZ: a single result on its predefined split).
Best and second-best scores are \textbf{bold} and
\underline{underlined}, respectively.
}
\resizebox{\textwidth}{!}{%
\begin{tabular}{l|cccccccc}
\toprule
\multirow{2}{*}[-0.6ex]{\textbf{Variant}}
& \multicolumn{2}{c}{\textbf{P300}}
& \multicolumn{2}{c}{\textbf{Motor Imagery}}
& \multicolumn{2}{c}{\textbf{Motor Execution}}
& \multicolumn{1}{c}{\textbf{Speech}}
& \multicolumn{1}{c}{\textbf{Seizure}} \\
\cmidrule(lr){2-3}
\cmidrule(lr){4-5}
\cmidrule(lr){6-7}
\cmidrule(lr){8-8}
\cmidrule(lr){9-9}
& BI2014b & BI2013a
& BNCI2014 & BNCI2015
& BNCI2020 & HDOP
& Broderick & TUSZ \\
\midrule

M1
& $43.03_{\pm 1.75}$ & $52.77_{\pm 1.22}$
& $43.11_{\pm 6.70}$ & $65.17_{\pm 5.33}$
& $39.49_{\pm 3.13}$ & $48.12_{\pm 3.84}$
& $38.02_{\pm 0.43}$ & $\underline{21.32}$ \\

M2
& $43.88_{\pm 2.46}$ & $53.46_{\pm 0.99}$
& $43.38_{\pm 5.52}$ & $\underline{65.67}_{\pm 4.30}$
& $\underline{40.92}_{\pm 4.04}$ & $48.87_{\pm 5.05}$
& $\underline{38.73}_{\pm 0.57}$ & $18.96$ \\

M3
& $\underline{44.54}_{\pm 2.32}$ & $\underline{53.89}_{\pm 0.86}$
& $\underline{43.72}_{\pm 7.79}$ & $65.44_{\pm 3.02}$
& $40.57_{\pm 7.63}$ & $\underline{48.97}_{\pm 3.17}$
& $38.71_{\pm 0.43}$ & $19.69$ \\

M4
& $43.46_{\pm 2.43}$ & $52.44_{\pm 1.22}$
& $42.39_{\pm 7.18}$ & $64.93_{\pm 4.11}$
& $40.49_{\pm 3.06}$ & $47.96_{\pm 3.22}$
& $38.12_{\pm 0.62}$ & $16.30$ \\

M5
& $42.83_{\pm 2.08}$ & $52.25_{\pm 1.61}$
& $42.83_{\pm 5.19}$ & $64.23_{\pm 4.53}$
& $28.67_{\pm 1.29}$ & $46.77_{\pm 3.38}$
& $37.87_{\pm 0.66}$ & $7.98$ \\

\midrule
 \rowcolor{SkyBlue!30} \textbf{Full}
& $\textbf{45.23}_{\pm 1.22}$
&  $\textbf{56.50}_{\pm 0.97}$
&  $\textbf{45.55}_{\pm 4.96}$
&  $\textbf{67.77}_{\pm 3.99}$
&  $\textbf{41.85}_{\pm 3.83}$
&  $\textbf{52.91}_{\pm 3.33}$
&  $\textbf{38.76}_{\pm 0.57}$
& $\textbf{25.06}$ \\
\bottomrule
\end{tabular}%
}
\label{loss_abl}
\end{table}

\subsection{Main Results}

As shown in Table~\ref{llava-main-result}, NeuroDETR ranks first
on most datasets in $F1_{\mathrm{sample}}$.
It achieves an average of 57.50, outperforming DANCE, the strongest
baseline on average, by 3.32 percentage points and demonstrating
consistently strong sample-wise discrimination.

For $F1_{\mathrm{event}}$, NeuroDETR achieves the best performance
on all eight datasets, averaging 46.70 and surpassing DANCE
by 4.29 percentage points.
Its improvement over the strongest baseline reaches 8.99 percentage
points on BNCI2015, highlighting its effectiveness in jointly
detecting, classifying, and localizing EEG events.

Some baselines achieve competitive sample-wise classification,
yet this advantage does not consistently translate into better
event detection.
NeuroDETR achieves the best average performance on both metrics,
combining strong sample-wise classification with effective
event-instance recovery.
Its consistent gains across P300, motor imagery, motor execution,
speech, and seizure tasks support its broad applicability across
EEG datasets and paradigms.

\begin{table}[t]
\centering
\renewcommand{\arraystretch}{1.1}

\caption{
Quantization and FOCAL ablation:
$F1_{\mathrm{event}}$ (\%, five-fold mean $\pm$ SD;
TUSZ: a single result).
}
\label{tab:quantization_guidance}

\resizebox{\textwidth}{!}{%
\begin{tabular}{cc|cccccccc}
\toprule
\multicolumn{2}{c|}{\textbf{Configuration}}
& \multicolumn{2}{c}{\textbf{P300}}
& \multicolumn{2}{c}{\textbf{Motor Imagery}}
& \multicolumn{2}{c}{\textbf{Motor Execution}}
& \textbf{Speech}
& \textbf{Seizure} \\

\cmidrule(lr){1-2}
\cmidrule(lr){3-4}
\cmidrule(lr){5-6}
\cmidrule(lr){7-8}
\cmidrule(lr){9-9}
\cmidrule(lr){10-10}

\textbf{Quantization}
& \textbf{FOCAL}
& BI2014b & BI2013a
& BNCI2014 & BNCI2015
& BNCI2020 & HDOP
& Broderick & TUSZ \\

\midrule

\ding{55} & \ding{55}
& $42.02_{\pm 1.38}$
& $50.64_{\pm 0.65}$
& $39.44_{\pm 3.27}$
& $63.05_{\pm 3.50}$
& $36.73_{\pm 3.61}$
& $46.34_{\pm 2.79}$
& $37.15_{\pm 0.30}$
& $12.24$ \\

\ding{51} & \ding{55}
& $43.50_{\pm 2.74}$
& $51.11_{\pm 1.32}$
& $41.23_{\pm 6.30}$
& $64.15_{\pm 5.11}$
& $37.28_{\pm 4.54}$
& $46.50_{\pm 1.43}$
& $\underline{37.98}_{\pm 0.44}$
& $16.32$ \\

\ding{55} & \ding{51}
& $\underline{43.98}_{\pm 2.64}$
& $\underline{51.99}_{\pm 1.33}$
& $\underline{42.76}_{\pm 5.88}$
& $\underline{65.34}_{\pm 2.88}$
& $\underline{37.79}_{\pm 3.70}$
& $\underline{47.98}_{\pm 3.04}$
& $37.94_{\pm 0.51}$
& $\underline{20.14}$ \\

\midrule

\rowcolor{SkyBlue!30}
\ding{51} & \ding{51}
& $\mathbf{45.23}_{\pm 1.22}$
& $\mathbf{56.50}_{\pm 0.97}$
& $\mathbf{45.55}_{\pm 4.96}$
& $\mathbf{67.77}_{\pm 3.99}$
& $\mathbf{41.85}_{\pm 3.83}$
& $\mathbf{52.91}_{\pm 3.33}$
& $\mathbf{38.76}_{\pm 0.57}$
& $\mathbf{25.06}$ \\

\bottomrule
\end{tabular}%
}
\end{table}

\subsection{Ablation Studies}

We evaluate the contributions of NeuroDETR's core components
and examine the relationship between spatiotemporal quantization
and event guidance.

\paragraph{Ablation of Core Components.}
We compare five variants with the full model.
M1 and M2 replace the temporal and spatial quantized representations
with continuous features, respectively.
M3 retains only the finest temporal scale while preserving
the memory grid and fusion output length.
M4 retains only fine electrode clusters, removing higher spatial
levels and cross-level residual connections.
M5 removes FOCAL and event-presence supervision, feeds backbone
features directly into temporal tokenization, and replaces
event-guided spatial pooling with adaptive average pooling.

Table~\ref{loss_abl} shows that the full model outperforms all five
variants in event F1 on each of the eight datasets.
The largest drops under M5 occur on BNCI2020, HDOP, and TUSZ,
reaching 13.18, 6.14, and 17.08 points, respectively.
These datasets have event occupancies of 27.03\%, 6.15\%,
and 3.18\%, respectively.
On Broderick, M5 reduces event F1 by 0.89 points.
These results show that the combined contribution of
FOCAL-associated processing and supervision varies across datasets,
with larger gains in these low-occupancy settings.
M4 also reduces event F1 by 4.95 points on HDOP and 8.76 points
on TUSZ, supporting the contribution of the multilevel spatial
representation and its cross-level residual connections.

\paragraph{Quantization and Event Guidance.}
We evaluate continuous and quantized representations with and
without event guidance in a $2\times2$ study.
Continuous variants bypass both quantizers while preserving
pooling, hierarchy, and fusion.
Disabling guidance replaces event-based routing with
event-independent window selection and uses uniform weights
for regional feature pooling.
The attention architecture, event-presence head, and its
supervision are retained across configurations.

For each dataset, let $F_{q,g}$ denote the reported event F1,
where $q,g\in\{0,1\}$ indicate whether quantization and event
guidance are enabled.
We measure their interaction as
\[
\Delta_{\mathrm{int}}
=(F_{1,1}-F_{0,1})-(F_{1,0}-F_{0,0}).
\]
A positive value indicates quantization provides a larger gain with event guidance than without it.

Table~\ref{tab:quantization_guidance} shows that the joint
configuration achieves the highest event F1 on all eight datasets.
Across the seven cross-validation datasets, the average
quantization gain increases from 0.91 points without guidance
to 2.97 points with guidance, yielding an average interaction
contrast of 2.06 points.
The interaction contrasts computed from the reported means
are largest on HDOP (4.77 points), BI2013a (4.04), and
BNCI2020 (3.51).
BI2014b shows a negative contrast ($-0.23$), while Broderick
is close to zero ($-0.01$).
These point estimates indicate that event guidance increases
the benefit of quantization on average, with interactions
that vary across datasets.

\subsection{Visualization of Decoded Events}

\begin{figure*}[t]
    \centering
    \begin{subfigure}[b]{0.49\textwidth}
        \centering
        \includegraphics[width=\linewidth]{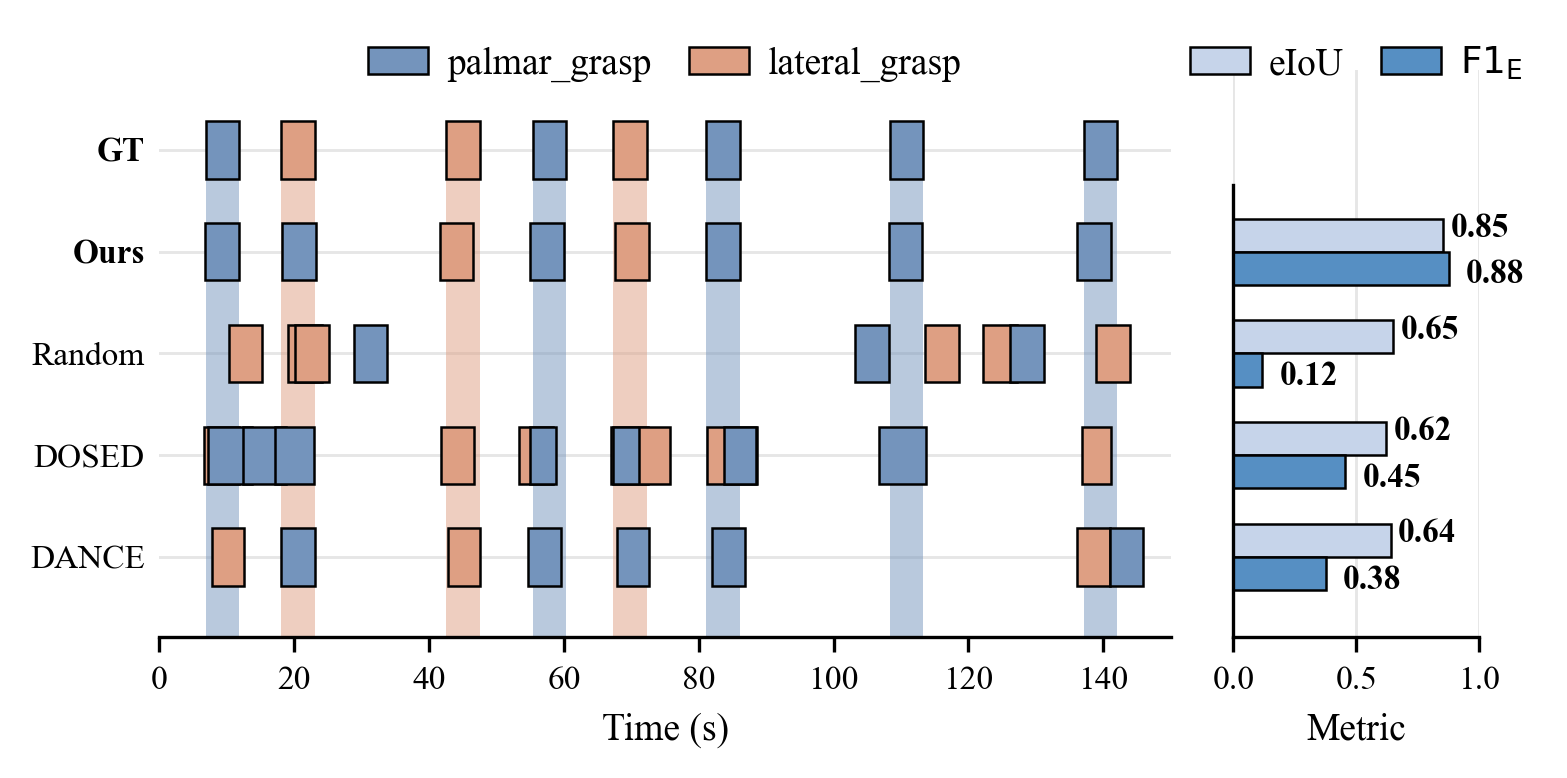}
        \caption{BNCI2020 (150\,s)}
        \label{fig:bnci2020}
    \end{subfigure}
    \hfill
    \begin{subfigure}[b]{0.49\textwidth}
        \centering
        \includegraphics[width=\linewidth]{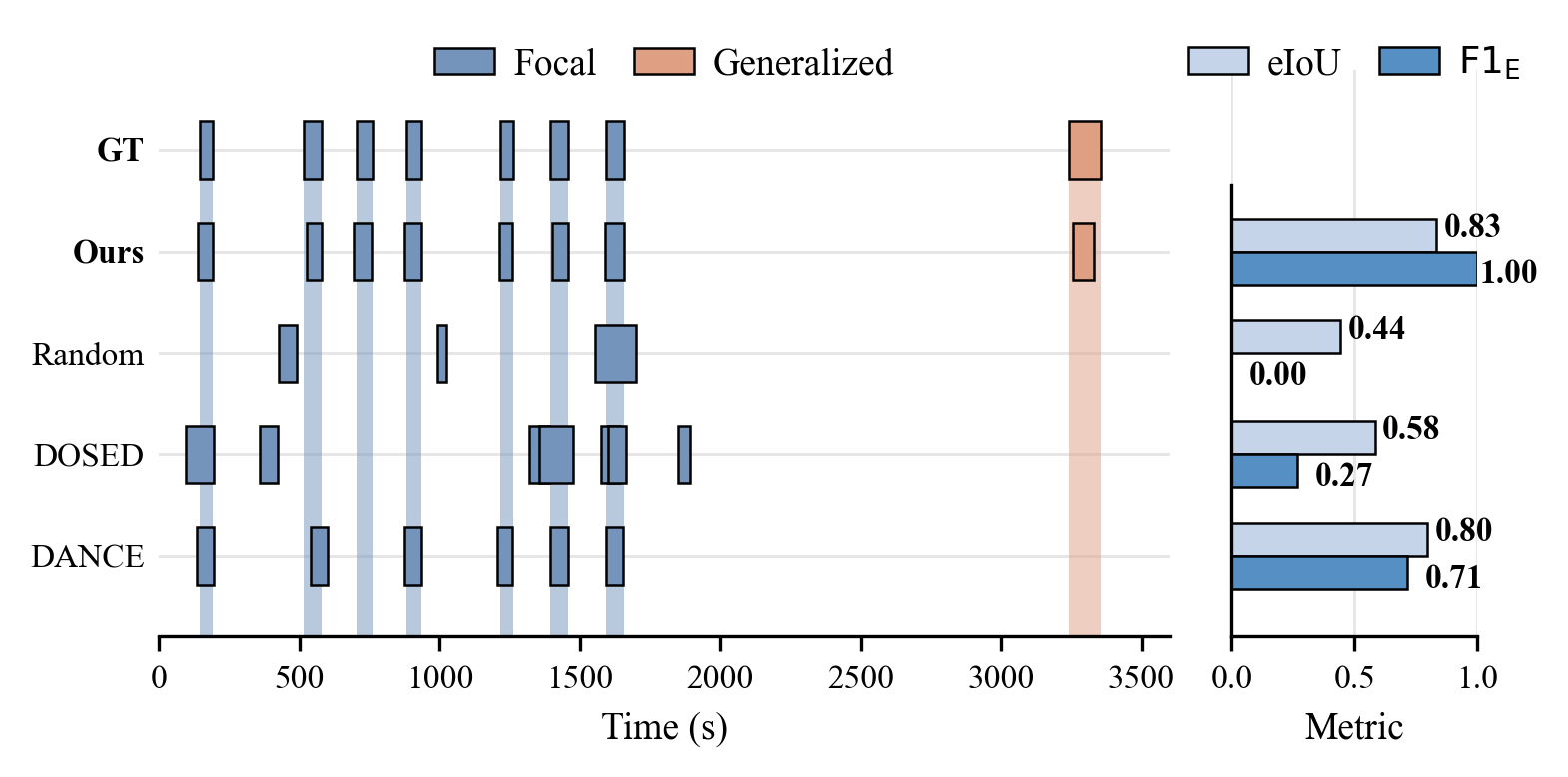}
        \caption{TUSZ (3600\,s)}
        \label{fig:tusz}
    \end{subfigure}
    \caption{
        Temporal decoding examples from (a) BNCI2020 and (b) TUSZ.
        Colored intervals indicate event classes and temporal extents;
        GT denotes ground truth.
        Side panels report metrics computed on the displayed segments:
        eIoU is the mean temporal IoU over overlapping, same-class
        one-to-one matches, and event-level F1 ($F1_{\mathrm{event}}$)
        is evaluated at an IoU threshold of 0.5.
        N/A indicates that no matches are available for computing eIoU.
    }
    \label{fig:two_datasets}
\end{figure*}

Figure~\ref{fig:two_datasets} compares NeuroDETR with Random,
DOSED, and DANCE on a 150\,s EEG segment from BNCI2020 and
a 3600\,s segment from TUSZ.
The aligned timelines show predicted event classes and temporal
boundaries alongside the ground-truth annotations, allowing
inspection of boundary deviations, missed events, and false positives.
The accompanying metrics summarize detection performance within
each displayed segment.

\section{Conclusion}
\label{sec:conclusion}

We presented NeuroDETR, a framework for asynchronous EEG event detection that jointly predicts event classes, onsets, and offsets from continuous recordings. Multiscale temporal VQ and Spatial-Pyramid VQ organize temporal dynamics and hierarchical spatial information into a compact event memory. FOCAL uses predicted event presence to guide temporal feature refinement and spatial pooling while preserving cross-window context. Across eight datasets spanning P300, motor imagery, motor execution, speech, and seizure detection, NeuroDETR achieves the highest event-level F1 among the compared methods, improving the average over DANCE by 4.29 percentage points. Ablation studies support the complementary contributions of discrete spatiotemporal representations, multiscale structure, and event-aware refinement. These results demonstrate the value of organizing event-relevant information across temporal scales and spatial regions for joint event classification and localization in continuous EEG.

\subsection*{AI Use Statement}

We used generative AI tools to assist with manuscript writing and language polishing, focusing on improving clarity and readability. We take full responsibility for the final content of this work, including all AI-assisted text. All AI-assisted text was reviewed and revised by the authors.

\subsection*{Reproducibility Statement}

The main text describes the architecture and training objectives
of NeuroDETR.
Dataset statistics, input window lengths, and per-GPU batch sizes
are summarized in Table~\ref{tab:dataset_statistics}.
Section~\ref{sec:implementation_details} specifies the preprocessing,
data splits, model hyperparameters, optimization settings,
random seed, and validation-based checkpoint selection procedure.

\bibliographystyle{unsrtnat}
\bibliography{iclr2027_conference}

@Article{tang2023flexible,
  author        = {Tang, Xin and Shen, Hao and Zhao, Siyuan and Li, Na and Liu, Jia},
  title         = {Flexible brain-computer interfaces},
  journal       = {Nat. Electron.},
  volume        = {6},
  number        = {2},
  pages         = {109--118},
  year          = {2023},
}

@Article{robinson2025application,
  author        = {Robinson, Jacob T. and Norman, Sumner L. and Angle, Matthew R. and Constandinou, Timothy G. and Denison, Timothy and Donoghue, John P. and Field, Ryan M. and others},
  title         = {An application-based taxonomy for brain-computer interfaces},
  journal       = {Nat. Biomed. Eng.},
  volume        = {9},
  number        = {6},
  pages         = {789--791},
  year          = {2025},
}

@Article{defossez2023decoding,
  author        = {D{\'e}fossez, Alexandre and Caucheteux, Charlotte and Rapin, J{\'e}r{\'e}my and Kabeli, Ori and King, Jean-R{\'e}mi},
  title         = {Decoding speech perception from non-invasive brain recordings},
  journal       = {Nat. Mach. Intell.},
  volume        = {5},
  number        = {10},
  pages         = {1097--1107},
  year          = {2023},
}

@Article{wu2024motor,
  author        = {Wu, Huanyu and Li, Siyang and Wu, Dongrui},
  title         = {Motor imagery classification for asynchronous {EEG}-based brain--computer interfaces},
  journal       = {IEEE Trans. Neural Syst. Rehabil. Eng.},
  volume        = {32},
  pages         = {527--536},
  year          = {2024},
}

@InProceedings{jaegle2021perceiver,
  author        = {Jaegle, Andrew and Gimeno, Felix and Brock, Andy and Vinyals, Oriol and Zisserman, Andrew and Carreira, Joao},
  title         = {{Perceiver}: General perception with iterative attention},
  booktitle     = {{P}roc. ICML},
  pages         = {4651--4664},
  year          = {2021},
}

@Article{zhang2008asynchronous,
  author        = {Zhang, Haihong and Guan, Cuntai and Wang, Chuanchu},
  title         = {Asynchronous {P300}-based brain--computer interfaces: A computational approach with statistical models},
  journal       = {IEEE Trans. Biomed. Eng.},
  volume        = {55},
  number        = {6},
  pages         = {1754--1763},
  year          = {2008},
}

@Article{lopes2019online,
  author        = {Lopes-Dias, Catarina and Sburlea, Andreea I. and M{\"u}ller-Putz, Gernot R.},
  title         = {Online asynchronous decoding of error-related potentials during the continuous control of a robot},
  journal       = {Sci. Rep.},
  volume        = {9},
  number        = {1},
  pages         = {17596},
  year          = {2019},
}

@Article{chambon2019dosed,
  author        = {Chambon, Stanislas and Thorey, Valentin and Arnal, Pierrick J. and Mignot, Emmanuel and Gramfort, Alexandre},
  title         = {{DOSED}: A deep learning approach to detect multiple sleep micro-events in {EEG} signal},
  journal       = {J. Neurosci. Methods},
  volume        = {321},
  pages         = {64--78},
  year          = {2019},
}

@InProceedings{zhang2022actionformer,
  author        = {Zhang, Chen-Lin and Wu, Jianxin and Li, Yin},
  title         = {{ActionFormer}: Localizing moments of actions with transformers},
  booktitle     = {{P}roc. ECCV},
  pages         = {492--510},
  year          = {2022},
}

@InProceedings{yang2024dyfadet,
  author        = {Yang, Le and Zheng, Ziwei and Han, Yizeng and Cheng, Hao and Song, Shiji and Huang, Gao and Li, Fan},
  title         = {{DyFADet}: Dynamic feature aggregation for temporal action detection},
  booktitle     = {{P}roc. ECCV},
  pages         = {305--322},
  year          = {2024},
}

@InProceedings{wolf2022deep,
  author        = {Wolf, Lukas and Kastrati, Ard and P{\l}omecka, Martyna Beata and Li, Jie-Ming and Klebe, Dustin and Veicht, Alexander and Wattenhofer, Roger and others},
  title         = {A deep learning approach for the segmentation of electroencephalography data in eye tracking applications},
  booktitle     = {{P}roc. ICML},
  pages         = {23912--23932},
  year          = {2022},
}

@Article{ma2026adapting,
  author        = {Ma, Jun and Ruotsalo, Tuukka},
  title         = {Adapting frozen foundation models for montage-agnostic high-resolution {EEG} event segmentation},
  journal       = {J. Neural Eng.},
  volume        = {23},
  number        = {3},
  pages         = {036020},
  year          = {2026},
}

@Misc{levy2026dance,
  author        = {L{\'e}vy, Jarod and Banville, Hubert and Rapin, J{\'e}r{\'e}my and King, Jean-Remi and Moreau, Thomas and d'Ascoli, St{\'e}phane},
  title         = {{DANCE}: Detect and classify events in {EEG}},
  year          = {2026},
  archiveprefix = {arXiv},
  eprint        = {2605.10688},
  journal       = {arXiv preprint},
}

@InProceedings{gui2024vector,
  author        = {Gui, Haokun and Li, Xiucheng and Chen, Xinyang},
  title         = {Vector quantization pretraining for {EEG} time series with random projection and phase alignment},
  booktitle     = {{P}roc. ICML},
  pages         = {16731--16750},
  year          = {2024},
}

@InProceedings{jiang2025neurolm,
  author        = {Jiang, Wei-Bang and Wang, Yansen and Lu, Bao-Liang and Li, Dongsheng},
  title         = {{NeuroLM}: A universal multi-task foundation model for bridging the gap between language and {EEG} signals},
  booktitle     = {{P}roc. ICLR},
  pages         = {55436--55457},
  year          = {2025},
}

@InProceedings{pradeepkumar2026tokenizing,
  author        = {Pradeepkumar, Jathurshan and Piao, Xihao and Chen, Zheng and Sun, Jimeng},
  title         = {Tokenizing single-channel {EEG} with time-frequency motif learning},
  booktitle     = {{P}roc. ICLR},
  pages         = {36611--36637},
  year          = {2026},
}

@InProceedings{ma2025codebrain,
  author        = {Ma, Jingying and Wu, Feng and Lin, Qika and Xing, Yucheng and Liu, Chenyu and Jia, Ziyu and Feng, Mengling},
  title         = {{CodeBrain}: Bridging decoupled tokenizer and multi-scale architecture for {EEG} foundation model},
  booktitle     = {{P}roc. ICLR},
  pages         = {83526--83574},
  year          = {2026},
}

@Article{ding2022tsception,
  author        = {Ding, Yi and Robinson, Neethu and Zhang, Su and Zeng, Qiuhao and Guan, Cuntai},
  title         = {{TSception}: Capturing temporal dynamics and spatial asymmetry from {EEG} for emotion recognition},
  journal       = {IEEE Trans. Affect. Comput.},
  volume        = {14},
  number        = {3},
  pages         = {2238--2250},
  year          = {2023},
}

@Article{gao2024graph,
  author        = {Gao, Hongxiang and Wang, Xingyao and Chen, Zhenghua and Wu, Min and Cai, Zhipeng and Zhao, Lulu and Li, Jianqing and others},
  title         = {Graph convolutional network with connectivity uncertainty for {EEG}-based emotion recognition},
  journal       = {IEEE J. Biomed. Health Inform.},
  volume        = {28},
  number        = {10},
  pages         = {5917--5928},
  year          = {2024},
}

@Article{liu2024empt,
  author        = {Liu, Ming and Liu, Yanbing and Shi, Weiyou and Lou, Yitai and Sun, Yuan and Meng, Qi and Wang, Dezheng and others},
  title         = {{EMPT}: A sparsity transformer for {EEG}-based motor imagery recognition},
  journal       = {Front. Neurosci.},
  volume        = {18},
  pages         = {1366294},
  year          = {2024},
}

@Article{miao2025satrans,
  author        = {Miao, Tianhua and Sha, Liansen and Huang, Kun and Li, Yongbin and Liu, Bin},
  title         = {{SATrans-Net}: Sparse attention transformer for {EEG}-based motor imagery decoding},
  journal       = {Sci. Rep.},
  volume        = {16},
  pages         = {1192},
  year          = {2026},
}

@InProceedings{carion2020end,
  author        = {Carion, Nicolas and Massa, Francisco and Synnaeve, Gabriel and Usunier, Nicolas and Kirillov, Alexander and Zagoruyko, Sergey},
  title         = {End-to-end object detection with transformers},
  booktitle     = {{P}roc. ECCV},
  pages         = {213--229},
  year          = {2020},
}

@Article{seeuws2024avoiding,
  author        = {Seeuws, Nick and De Vos, Maarten and Bertrand, Alexander},
  title         = {Avoiding post-processing with event-based detection in biomedical signals},
  journal       = {IEEE Trans. Biomed. Eng.},
  volume        = {71},
  number        = {8},
  pages         = {2442--2453},
  year          = {2024},
}

@InProceedings{congedo2011brain,
  author        = {Congedo, Marco and Goyat, Matthieu and Tarrin, Nicolas and Ionescu, Gelu and Varnet, L{\'e}o and Rivet, Bertrand and Phlypo, Ronald and others},
  title         = {``{Brain Invaders}'': A prototype of an open-source {P300}-based video game working with the {OpenViBE} platform},
  booktitle     = {Int. BCI Conf.},
  pages         = {280--283},
  year          = {2011},
}

@Misc{korczowski_2019_3267302,
  author        = {Korczowski, Louis and Ostaschenko, Ekaterina and Andreev, Anton and Cattan, Gr{\'e}goire and Coelho Rodrigues, Pedro Luis and Gautheret, Violette and Congedo, Marco},
  title         = {{Brain Invaders} solo versus collaboration: Multi-user {P300}-based brain-computer interface dataset ({bi2014b})},
  year          = {2019},
  howpublished  = {Zenodo},
  doi           = {10.5281/zenodo.3267302},
}

@InProceedings{jiang2024large,
  author        = {Jiang, Wei-Bang and Zhao, Li-Ming and Lu, Bao-Liang},
  title         = {Large brain model for learning generic representations with tremendous {EEG} data in {BCI}},
  booktitle     = {{P}roc. ICLR},
  pages         = {16405--16426},
  year          = {2024},
}

@Misc{ho2019axial,
  author        = {Ho, Jonathan and Kalchbrenner, Nal and Weissenborn, Dirk and Salimans, Tim},
  title         = {Axial attention in multidimensional transformers},
  year          = {2019},
  archiveprefix = {arXiv},
  eprint        = {1912.12180},
  journal       = {arXiv preprint},
}

@Article{kuhn1955hungarian,
  author        = {Kuhn, H. W.},
  title         = {The {Hungarian} method for the assignment problem},
  journal       = {Nav. Res. Logist. Q.},
  volume        = {2},
  number        = {1--2},
  pages         = {83--97},
  year          = {1955},
}

@Misc{barmpas2025neurorvq,
  author        = {Barmpas, Konstantinos and Lee, Na and Chalatsis, Dimitrios and Raftery, William and Panagakis, Yannis and Adamos, Dimitrios A. and Laskaris, Nikolaos and others},
  title         = {{NeuroRVQ}: Multi-scale biosignal tokenization for generative foundation models},
  year          = {2026},
  archiveprefix = {arXiv},
  eprint        = {2510.13068v4},
  journal       = {arXiv preprint},
}

@InProceedings{yang2026thd,
  author        = {Yang, Wenchao and Yan, Weidong and Liu, Wenkang and Ma, Yulan and Li, Yang},
  title         = {{THD-BAR}: Topology hierarchical derived brain autoregressive modeling for {EEG} generic representations},
  booktitle     = {{P}roc. NeurIPS},
  volume        = {38},
  pages         = {150290--150314},
  year          = {2025},
}

@InProceedings{van2017neural,
  author        = {van den Oord, Aaron and Vinyals, Oriol and Kavukcuoglu, Koray},
  title         = {Neural discrete representation learning},
  booktitle     = {{P}roc. NeurIPS},
  volume        = {30},
  pages         = {6306--6315},
  year          = {2017},
}

@Article{broderick2018electrophysiological,
  author        = {Broderick, Michael P. and Anderson, Andrew J. and Di Liberto, Giovanni M. and Crosse, Michael J. and Lalor, Edmund C.},
  title         = {Electrophysiological correlates of semantic dissimilarity reflect the comprehension of natural, narrative speech},
  journal       = {Curr. Biol.},
  volume        = {28},
  number        = {5},
  pages         = {803--809.e3},
  year          = {2018},
}

@Misc{broderick2019data,
  author        = {Broderick, Michael P. and Anderson, Andrew J. and Di Liberto, Giovanni M. and Crosse, Michael J. and Lalor, Edmund C.},
  title         = {Data from: Electrophysiological correlates of semantic dissimilarity reflect the comprehension of natural, narrative speech},
  year          = {2019},
  howpublished  = {Dryad},
  doi           = {10.5061/dryad.070jc},
}

@Article{tangermann2012review,
  author        = {Tangermann, Michael and M{\"u}ller, Klaus-Robert and Aertsen, Ad and Birbaumer, Niels and Braun, Christoph and Brunner, Clemens and Leeb, Robert and others},
  title         = {Review of the {BCI Competition IV}},
  journal       = {Front. Neurosci.},
  volume        = {6},
  pages         = {55},
  year          = {2012},
}

@Article{faller2012autocalibration,
  author        = {Faller, Josef and Vidaurre, Carmen and Solis-Escalante, Teodoro and Neuper, Christa and Scherer, Reinhold},
  title         = {Autocalibration and recurrent adaptation: Towards a plug and play online {ERD-BCI}},
  journal       = {IEEE Trans. Neural Syst. Rehabil. Eng.},
  volume        = {20},
  number        = {3},
  pages         = {313--319},
  year          = {2012},
}

@Article{schwarz2020analyzing,
  author        = {Schwarz, Andreas and Escolano, Carlos and Montesano, Luis and M{\"u}ller-Putz, Gernot R.},
  title         = {Analyzing and decoding natural reach-and-grasp actions using gel, water and dry {EEG} systems},
  journal       = {Front. Neurosci.},
  volume        = {14},
  pages         = {849},
  year          = {2020},
}

@Article{lomele2026high,
  author        = {Lomele, G. and Lencioni, T. and D'Ambrosio, S. and Comanducci, A. and Lucchetti, F. and Marzegan, A. and Derchi, C. and others},
  title         = {High-density {EEG} and multi-muscle {EMG} dataset during object prehension with a sensorized grasping box in humans},
  journal       = {Sci. Data},
  volume        = {13},
  number        = {1},
  pages         = {932},
  year          = {2026},
}

@Article{shah2018temple,
  author        = {Shah, Vinit and von Weltin, Eva and Lopez, Silvia and McHugh, James Riley and Veloso, Lillian and Golmohammadi, Meysam and Obeid, Iyad and others},
  title         = {The {Temple University Hospital Seizure Detection Corpus}},
  journal       = {Front. Neuroinform.},
  volume        = {12},
  pages         = {83},
  year          = {2018},
}

@InProceedings{duan2023dewave,
  author        = {Duan, Yiqun and Zhou, Jinzhao and Wang, Zhen and Wang, Yu-Kai and Lin, Chin-teng},
  title         = {{DeWave}: Discrete encoding of {EEG} waves for {EEG} to text translation},
  booktitle     = {{P}roc. NeurIPS},
  volume        = {36},
  pages         = {9907--9918},
  year          = {2023},
}

@InProceedings{rezatofighi2019generalized,
  author        = {Rezatofighi, Hamid and Tsoi, Nathan and Gwak, JunYoung and Sadeghian, Amir and Reid, Ian and Savarese, Silvio},
  title         = {Generalized intersection over union: A metric and a loss for bounding box regression},
  booktitle     = {{P}roc. CVPR},
  pages         = {658--666},
  year          = {2019},
}

@InProceedings{lin2017focal,
  author        = {Lin, Tsung-Yi and Goyal, Priya and Girshick, Ross and He, Kaiming and Doll{\'a}r, Piotr},
  title         = {Focal loss for dense object detection},
  booktitle     = {{P}roc. ICCV},
  pages         = {2980--2988},
  year          = {2017},
}

\newpage
\appendix
\textbf{Appendix}

\section{Additional Results}

\subsection{Event-Level F1 at Different Temporal IoU Thresholds}

\begin{figure*}[!htbp]
    \centering
    \includegraphics[width=\textwidth]{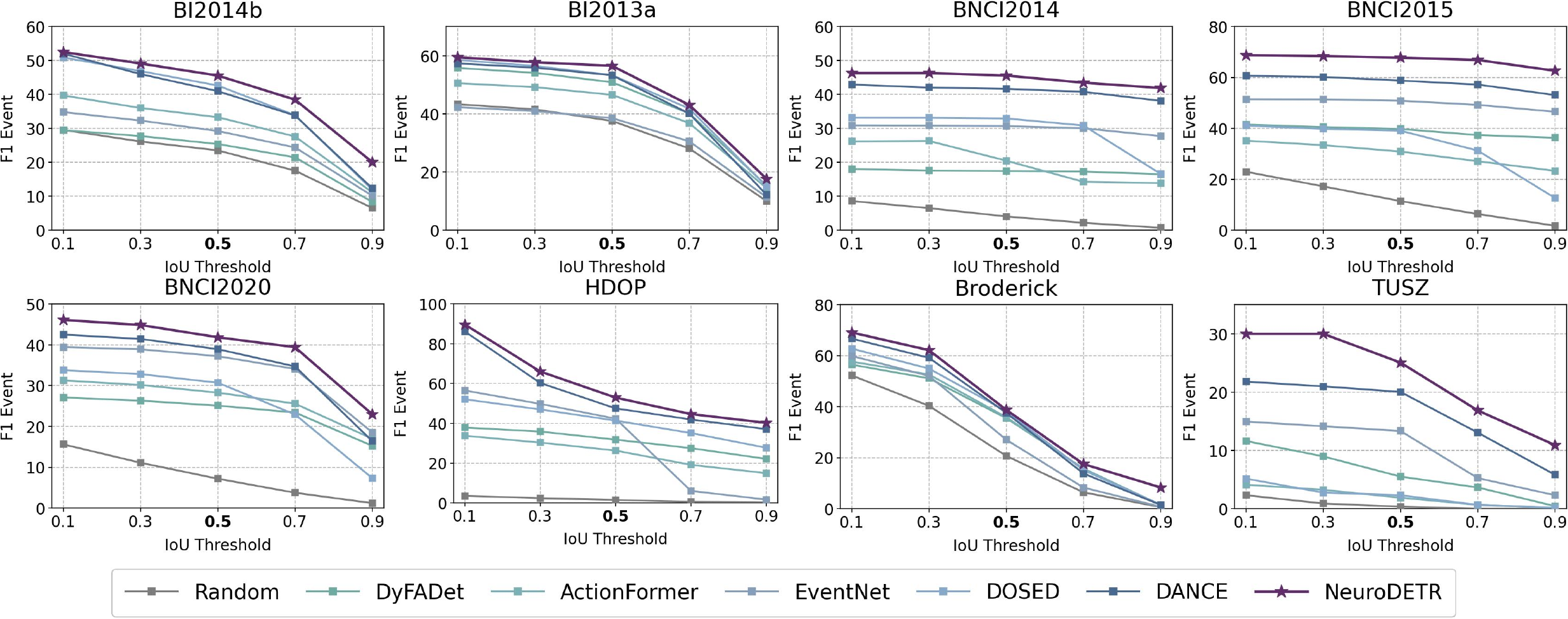}
    \caption{Event-level F1 of NeuroDETR and baseline methods at different temporal IoU thresholds. Higher thresholds impose stricter temporal localization requirements.}
    \label{fig:iou}
\end{figure*}

As shown in Figure~\ref{fig:iou}, we compare NeuroDETR with the baselines in $F1_{\mathrm{event}}$ at temporal IoU thresholds $\tau_{\mathrm{IoU}}\in\{0.1,0.3,0.5,0.7,0.9\}$ to assess detection performance under increasingly strict localization requirements.
Lower thresholds allow greater boundary deviations, whereas higher thresholds require closer alignment between predicted and annotated intervals.

NeuroDETR achieves the highest event-level F1 among the compared methods at every evaluated IoU threshold, demonstrating its effectiveness in joint event classification and temporal localization even under stringent boundary-matching requirements.

\subsection{Detection Error Analysis}
To distinguish missed events from incorrect detections, we evaluate the event-level false discovery rate (FDR) and false negative rate (FNR). Predictions are matched one-to-one to ground-truth events in descending confidence order, with a true positive requiring the correct class and temporal IoU $>0.5$. The metrics are defined as
\begin{equation}
    \mathrm{FDR}=\frac{\mathrm{FP}}{\mathrm{TP}+\mathrm{FP}}\times100\%,
    \qquad
    \mathrm{FNR}=\frac{\mathrm{FN}}{\mathrm{TP}+\mathrm{FN}}\times100\%.
    \label{eq:fdr_fnr}
\end{equation}
Here, TP counts correct matches, FP counts incorrect predictions, and FN counts ground-truth events without a correct match. FDR measures the proportion of incorrect detections among predictions, while FNR measures the proportion of missed events among annotations.

\begin{figure*}[t]
\centering
\includegraphics[width=\textwidth]{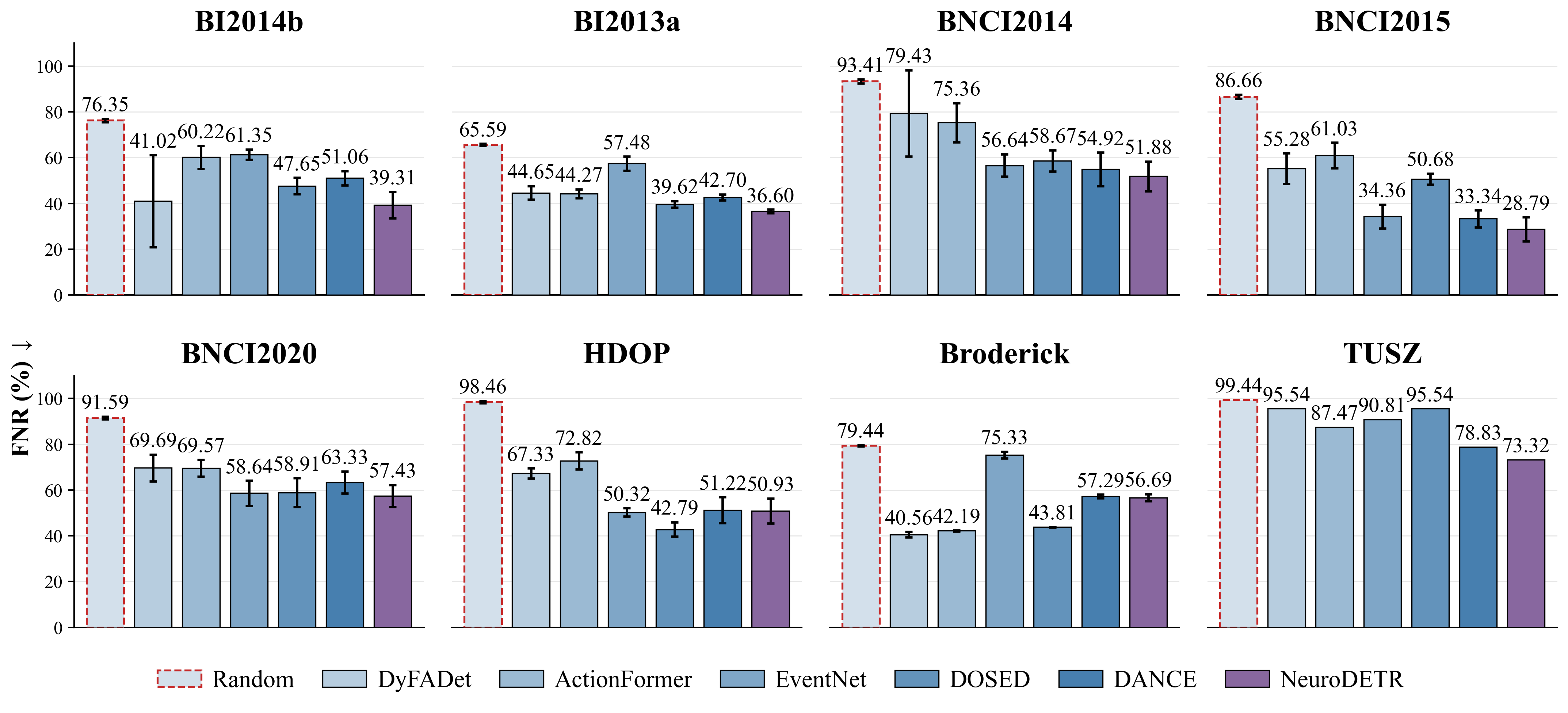}
\caption{Event-level false negative rate (FNR) across eight datasets. Lower values indicate fewer missed events. NeuroDETR achieves the lowest FNR on six datasets. Error bars denote standard deviations and are omitted for TUSZ, where standard deviations are unavailable.}
\label{fig:fnr}
\end{figure*}

\begin{figure*}[t]
\centering
\includegraphics[width=\textwidth]{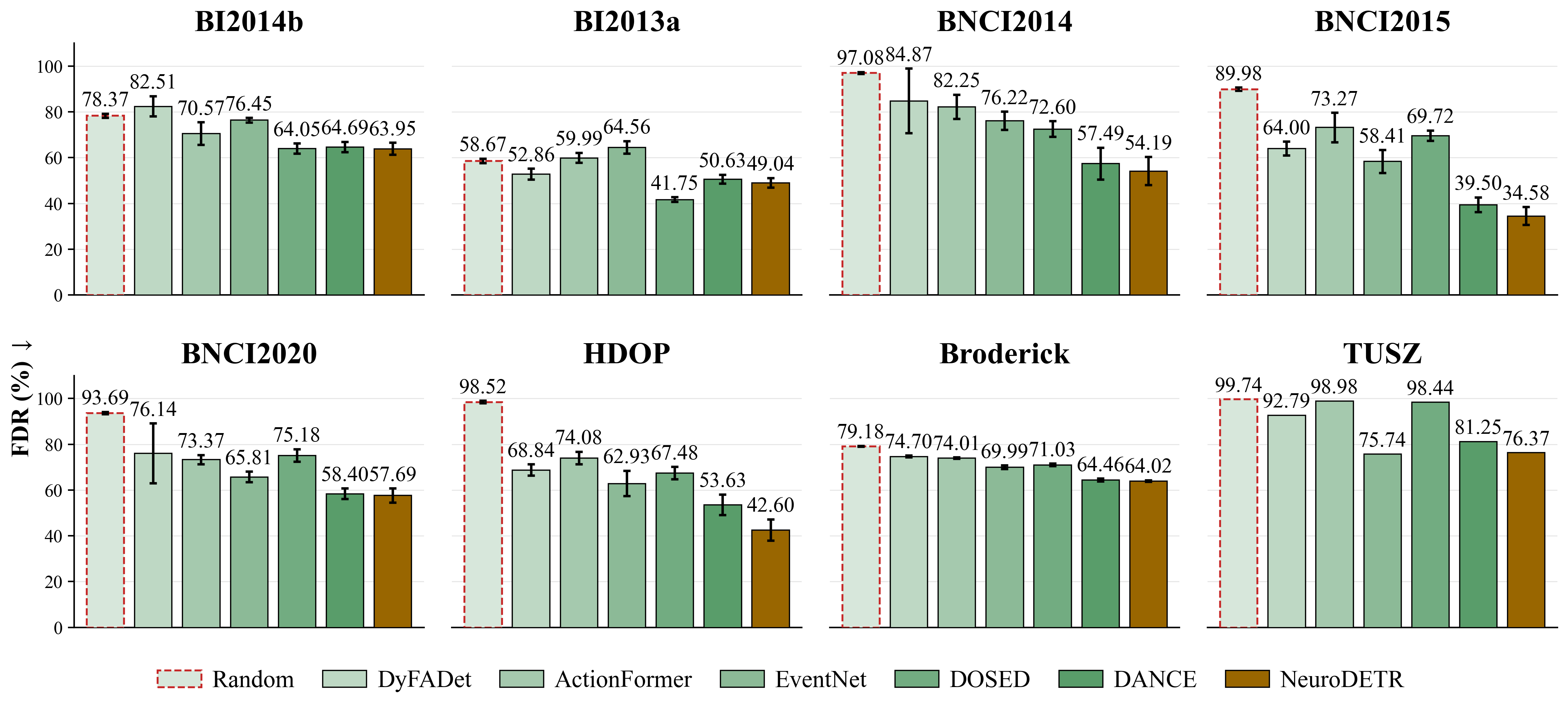}
\caption{Event-level false discovery rate (FDR) across eight datasets. Lower values indicate a smaller proportion of incorrect detections among predictions. NeuroDETR achieves the lowest FDR on six datasets. Error bars denote standard deviations and are omitted for TUSZ, where standard deviations are unavailable.}
\label{fig:fdr}
\end{figure*}

As shown in Figures~\ref{fig:fnr} and~\ref{fig:fdr}, NeuroDETR achieves the lowest FDR and FNR on six of the eight datasets each, while attaining the highest $\mathrm{F1}_{\mathrm{event}}$ on all eight. Where competing methods achieve a lower individual error rate, this advantage is accompanied by a higher complementary error rate: lower FDR coincides with more missed events, whereas lower FNR coincides with a larger proportion of incorrect detections. Such single-metric advantages therefore do not translate into better overall detection performance. NeuroDETR achieves the highest $\mathrm{F1}_{\mathrm{event}}$ across all eight datasets, indicating superior overall event detection performance.

\subsection{Sensitivity to the Number of Event Queries}
\begin{table*}[t]
    \centering
    \footnotesize
    \renewcommand{\arraystretch}{.9}
    \caption{
    Sensitivity to decoder query count $N_e$.
    For each dataset, the upper and lower rows report
    $N_e$ and $F1_{\mathrm{event}}$, respectively.
    \textbf{Bold} indicates the selected query count
    and its corresponding score.
    }
    \label{tab:query_sensitivity}

    \begin{tabular*}{0.5\textwidth}{@{\extracolsep{\fill}}lrrrrr@{}}
        \toprule
        Dataset & \multicolumn{5}{c}{Decoder queries / Event F1} \\
        \midrule
        BI2014b
        & 50 & 100 & \textbf{150} & 200 & 300 \\
        & 44.00 & 45.13 & \textbf{45.23} & 44.20 & 43.20 \\
        \midrule
        BI2013a
        & 50 & 100 & \textbf{150} & 200 & 300 \\
        & 55.65 & 52.71 & \textbf{56.50} & 52.46 & 54.28 \\
        \midrule
        BNCI2014
        & 1 & 3 & \textbf{5} & 10 & 20 \\
        & 42.55 & 43.10 & \textbf{45.55} & 45.44 & 42.77 \\
        \midrule
        BNCI2015
        & 10 & 20 & \textbf{40} & 60 & 80 \\
        & 66.85 & 67.44 & \textbf{67.77} & 67.49 & 64.12 \\
        \midrule
        BNCI2020
        & 10 & 20 & \textbf{40} & 60 & 80 \\
        & 40.03 & 41.32 & \textbf{41.85} & 40.99 & 39.80 \\
        \midrule
        HDOP
        & 4 & 8 & \textbf{16} & 24 & 32 \\
        & 51.52 & 51.78 & \textbf{52.91} & 41.17 & 51.94 \\
        \midrule
        Broderick
        & 50 & 100 & \textbf{150} & 200 & 400 \\
        & 38.65 & 38.73 & \textbf{38.76} & 38.71 & 37.05 \\
        \midrule
        TUSZ
        & 2 & 4 & \textbf{8} & 16 & 32 \\
        & 23.65 & 24.09 & \textbf{25.06} & 25.00 & 24.18 \\
        \bottomrule
    \end{tabular*}
\end{table*}

The number of decoder queries $N_e$ bounds the number of
event predictions within an input window.
Table~\ref{tab:query_sensitivity} reports event F1 at five
query counts for each dataset to assess sensitivity to
this prediction capacity.

Several datasets show small differences over a range
of query counts.
On Broderick, event F1 ranges from 38.65 to 38.76 for
$N_e$ between 50 and 200, before decreasing to 37.05
at 400 queries.
BNCI2014 yields scores of 45.55 and 45.44 with 5 and
10 queries, respectively, while TUSZ achieves 25.06
and 25.00 with 8 and 16 queries.

Increasing the query count can also produce substantial,
non-monotonic changes.
On BNCI2015, event F1 decreases from 67.77 to 64.12
when $N_e$ increases from 40 to 80.
HDOP shows a sharper change: increasing $N_e$ from 16
to 24 reduces event F1 from 52.91 to 41.17, followed
by a recovery to 51.94 at 32 queries.

Overall, query-count sensitivity depends on the dataset:
some query ranges yield similar scores, while individual
settings can produce substantial drops.
Larger query budgets do not consistently improve event
detection.
These findings motivate selecting $N_e$ separately for
each dataset using validation performance.

\section{Hierarchical Brain-Region Grouping}
\label{app:brain_region_grouping}

NeuroDETR uses a fixed three-level hierarchy of scalp-electrode
groups: one global group ($S_1$), five coarse regions ($S_2$),
and twelve fine clusters ($S_3$).
The global group aggregates all electrodes assigned to the
hierarchy, while the coarse-to-fine organization is summarized
in Table~\ref{tab:brain_region_groups}.

\begin{table}[htbp!]
\centering
\caption{Coarse-to-fine spatial grouping.}
\label{tab:brain_region_groups}
\resizebox{0.6\linewidth}{!}{%
\begin{tabular}{llc}
\toprule
\textbf{Coarse region ($S_2$)}
& \textbf{Fine clusters ($S_3$)}
& \textbf{Count} \\
\midrule
Frontal
& Left, midline, and right frontal
& 3 \\
Temporal
& Left and right temporal
& 2 \\
Central
& Left, midline, and right central
& 3 \\
Parietal
& Left, midline, and right parietal
& 3 \\
Occipital
& Undivided occipital group
& 1 \\
\midrule
Total & & 12 \\
\bottomrule
\end{tabular}%
}
\end{table}

Each fine cluster belongs to one coarse region, and all coarse
regions share the global group as their parent.
The two temporal clusters are merged at the coarse level,
whereas the occipital group remains undivided.
This hierarchy provides a spatial prior over scalp electrodes
and is instantiated using the available mapped channels
of each dataset.

\section{Details of Event Supervision}
\label{app:training_objectives}

This appendix specifies the set-prediction loss and auxiliary
supervision in Eq.~\ref{eq:method_total_loss}.
For each of the $N_e$ queries, $\hat{\mathbf p}_j$ denotes the
class probabilities over $\{0,\ldots,K\}$ and
$\hat I_j=[\hat s_j,\hat e_j]$ the predicted interval.
Ground-truth event $i$ has class $c_i$ and interval
$I_i=[s_i,e_i]$, with $M$ events in the input window.
Class $0$ denotes background, and all interval boundaries are
normalized to the input-window duration.

\subsection{Matching and Set-Prediction Loss}

\paragraph{Hungarian matching.}
Following set-based detection~\citep{carion2020end},
Hungarian matching produces a one-to-one assignment $\mathcal M$
by minimizing the total matching cost, with pairwise cost
\begin{equation}
    \mathcal C_{ji}
    =-\lambda_{\mathrm{mcls}}\log\hat p_j(c_i)
    +1-\operatorname{IoU}(\hat I_j,I_i).
    \label{eq:app_matching_cost}
\end{equation}
The assignment determines the classification targets and the
query--event pairs used for localization supervision.

\paragraph{Endpoint regression.}
For matched queries, the endpoint loss penalizes absolute errors
in normalized onset and offset predictions:
\begin{equation}
    \mathcal L_{\ell_1}
    =\frac{1}{N_e}
    \sum_{(j,i)\in\mathcal M}
    \left(
        |\hat s_j-s_i|+|\hat e_j-e_i|
    \right).
    \label{eq:app_endpoint_loss}
\end{equation}

\paragraph{Event classification.}
Matched queries receive their assigned event labels, while
unmatched queries receive the background label.
Cross-entropy supervision is applied to all queries:
\begin{equation}
    c_j^\star=
    \begin{cases}
        c_i, & (j,i)\in\mathcal M,\\
        0, & \text{otherwise},
    \end{cases}
    \qquad
    \mathcal L_{\mathrm{cls}}
    =-\frac{1}{N_e}\sum_{j=1}^{N_e}
    \log\hat p_j(c_j^\star).
    \label{eq:app_classification_loss}
\end{equation}

\paragraph{Interval localization.}
Temporal generalized IoU~\citep{rezatofighi2019generalized}
complements endpoint regression with interval-level supervision.
For intervals $A$ and $B$, let $H(A,B)$ denote their smallest
enclosing interval and $|\cdot|$ denote temporal length.
We define
\begin{equation}
\begin{aligned}
    \operatorname{IoU}(A,B)
    &=\frac{|A\cap B|}{|A\cup B|},\\
    \operatorname{GIoU}(A,B)
    &=\operatorname{IoU}(A,B)
    -\frac{|H(A,B)|-|A\cup B|}{|H(A,B)|},\\
    \mathcal L_{\mathrm{GIoU}}
    &=\frac{1}{N_e}
    \sum_{(j,i)\in\mathcal M}
    \left[
        1-\operatorname{GIoU}(\hat I_j,I_i)
    \right].
\end{aligned}
    \label{eq:app_giou_loss}
\end{equation}
The enclosing-interval term provides localization supervision
even when the predicted and target intervals do not overlap.

Both localization losses are normalized by the full query count
$N_e$, and unmatched queries receive no localization gradients.
For windows without annotated events, all classification targets
are background and both localization losses are zero.
The set-prediction objective is
\begin{equation}
    \mathcal L_{\mathrm{det}}
    =\lambda_{\ell_1}\mathcal L_{\ell_1}
    +\lambda_{\mathrm{cls}}\mathcal L_{\mathrm{cls}}
    +\lambda_{\mathrm{giou}}\mathcal L_{\mathrm{GIoU}}.
    \label{eq:app_detection_objective}
\end{equation}

\subsection{Auxiliary Supervision}

We complement set prediction with full-resolution boundary
supervision, dense classification, and prediction consistency.

\paragraph{Boundary supervision.}
A linear head on the backbone features $\mathbf H$ predicts
class-agnostic onset and offset logits $b_t^u$, where
$u\in\{\mathrm{on},\mathrm{off}\}$.
Let $\mu_i^u$ denote the corresponding annotated boundary
in sample coordinates.
For windows containing events, the target heatmaps are
\begin{equation}
    Y_t^u
    =\max_{i=1,\ldots,M}
    \exp\!\left(
        -\frac{(t-\mu_i^u)^2}{2\sigma_{\mathrm{bnd}}^2}
    \right),
    \qquad t=1,\ldots,T,
    \label{eq:app_boundary_targets}
\end{equation}
where $\sigma_{\mathrm{bnd}}$ controls the Gaussian width.
For windows without events, $Y_t^u=0$ for both boundary types.
Sigmoid focal loss~\citep{lin2017focal} supervises the heatmaps
while downweighting easy non-boundary positions:
\begin{equation}
    \mathcal L_{\mathrm{bnd}}
    =\frac{1}{2T}
    \sum_{u\in\{\mathrm{on},\mathrm{off}\}}
    \sum_{t=1}^{T}
    \operatorname{FL}_{\mathrm{sigmoid}}(b_t^u,Y_t^u).
    \label{eq:app_boundary_loss}
\end{equation}

\paragraph{Dense classification.}
A classifier on the fused memory
$\mathbf Z$ produces class probabilities
$\mathbf P_{\mathrm{dense}}\in\mathbb R^{N_q\times(K+1)}$.
Let $y_r$ denote the annotation label at grid position $r$,
with $0$ indicating background.
The dense classification loss is
\begin{equation}
    \mathcal L_{\mathrm{dense}}
    =-\frac{1}{N_q}\sum_{r=1}^{N_q}
    \log P_{\mathrm{dense}}(r,y_r).
    \label{eq:app_dense_loss}
\end{equation}

\paragraph{Prediction consistency.}
We project query predictions onto the same $N_q$-position
grid as the dense predictions.
Let $a_{rj}\in\{0,1\}$ indicate whether position $r$ lies
within query $j$'s discretized predicted interval.
We accumulate class probabilities from all queries, including
the background class, and normalize the scores at covered
positions:
\begin{equation}
\begin{aligned}
    S_{r,c}
    &=\sum_{j=1}^{N_e}a_{rj}\hat p_j(c),\\
    P_{\mathrm{set}}(r,c)
    &=
    \begin{cases}
        \displaystyle
        \frac{S_{r,c}}{\sum_{k=0}^{K}S_{r,k}},
        & \sum_{k=0}^{K}S_{r,k}>0,\\[6pt]
        0, & \text{otherwise}.
    \end{cases}
\end{aligned}
    \label{eq:app_set_projection}
\end{equation}
Overlapping predictions contribute jointly, while uncovered
positions are assigned zero vectors.
Let $\widetilde P$ denote componentwise clamping for numerical
stability, without subsequent renormalization, and let
$\epsilon_{\mathrm{clamp}}>0$ denote its lower bound.
The KL-style consistency penalty is averaged over the full grid:
\begin{equation}
    \mathcal L_{\mathrm{cons}}
    =\frac{1}{N_q}\sum_{r=1}^{N_q}\sum_{c=0}^{K}
    \widetilde P_{\mathrm{dense}}(r,c)
    \log
    \frac{\widetilde P_{\mathrm{dense}}(r,c)}
         {\widetilde P_{\mathrm{set}}(r,c)}.
    \label{eq:app_consistency_loss}
\end{equation}

At covered positions, this penalty encourages agreement
between the dense and set-derived predictions.
At uncovered positions, all components of the set-derived
vector are clamped to $\epsilon_{\mathrm{clamp}}$.
When clamping leaves the dense probabilities unchanged,
the contribution at an uncovered position, before averaging
over the grid, reduces to
\[
    \sum_{c=0}^{K} P_{\mathrm{dense}}(r,c)
    \log\frac{P_{\mathrm{dense}}(r,c)}
              {\epsilon_{\mathrm{clamp}}}
    =
    \log\frac{1}{\epsilon_{\mathrm{clamp}}}
    - H\!\left(P_{\mathrm{dense}}(r,\cdot)\right),
\]
where $H$ denotes Shannon entropy.
This term therefore favors higher-entropy dense predictions
at uncovered positions, while $\mathcal L_{\mathrm{dense}}$
provides supervision from the annotation labels.

At covered positions, gradients propagate to both
classification heads through probability entries unaffected
by clamping.
At uncovered positions, the set-derived vector is constant,
so this term backpropagates only through the dense predictions.
The discrete interval masks block gradients to the predicted
boundaries through this projection.

The event-supervision objective combines the set-prediction
loss with the three auxiliary terms:
\begin{equation}
\begin{aligned}
    \mathcal L_{\mathrm{event}}
    ={}&\mathcal L_{\mathrm{det}}
    +\lambda_{\mathrm{bnd}}\mathcal L_{\mathrm{bnd}} +\lambda_{\mathrm{dense}}\mathcal L_{\mathrm{dense}}
    +\lambda_{\mathrm{cons}}\mathcal L_{\mathrm{cons}}.
\end{aligned}
    \label{eq:app_event_objective}
\end{equation}

\section{Datasets}
\label{sec:dataset_implementation}

\begin{table}[!htbp]
\centering
\caption{Dataset statistics and training configurations.}
\label{tab:dataset_statistics}
\begingroup
\renewcommand{\arraystretch}{1.3}
\resizebox{0.95\textwidth}{!}{%
\begin{tabular}{llccccccccc}
\toprule
Dataset & Domain & Subjects & Channels
& \shortstack{Event\\(h)}
& \shortstack{Non-event\\(h)}
& \shortstack{Mean event\\duration (s)}
& \shortstack{Event coverage\\(\%)}
& Split
& \shortstack{Window\\(s)}
& \shortstack{Batch size\\per GPU} \\
\midrule
Korczowski2014b & P300 & 38 & 32
& 1.0 & 1.3 & 1.0 & 44 & CV & 8 & 32 \\
Vaineau2018Brain & P300 & 24 & 16
& 4.4 & 1.1 & 1.0 & 80 & CV & 32 & 8 \\
Tangermann2012 & Motor imagery & 9 & 22
& 5.8 & 5.8 & 4.0 & 50 & CV & 8 & 32 \\
Faller2012 & Motor imagery & 12 & 13
& 7.8 & 8.9 & 5.0 & 47 & CV & 32 & 4 \\
BNCI2020 & Motor execution & 45 & 12/32/58
& 10 & 27 & 5.0 & 27 & CV & 32 & 4 \\
HDOP & Motor execution & 14 & 62
& 0.24 & 3.7 & 0.20 & 6.1 & CV & 9 & 8 \\
Broderick & Speech & 19 & 128
& 14 & 6.5 & 0.23 & 69 & CV & 8 & 32 \\
TUSZ (v2.0.6) & Seizure & 675 & 17--25
& 47 & $1.4\times10^{3}$ & 63 & 3.2 & Fixed & 256
& $4^{\dagger}$ \\
\bottomrule
\end{tabular}%
}

\begin{minipage}{\textwidth}
\footnotesize
$^{\dagger}$TUSZ uses eight GPUs with a batch size of four per GPU,
giving an effective batch size of 32 ($8 \times 4$).
\end{minipage}
\endgroup
\end{table}

Table~\ref{tab:dataset_statistics} summarizes the eight EEG datasets used
in our experiments, including their domains, subject and channel counts,
event and non-event recording times, mean event durations, event coverage,
data splits, and input window lengths. For each recording, event time is the temporal union of the
configured event annotations, and non-event time is the remaining recording
time. Summing these quantities across recordings gives $E$ and $N$, reported
in hours, with event coverage defined as $E/(E+N)\times100\%$.
Overlapping intervals contribute only once to $E$.

The datasets span P300, motor imagery, motor execution, speech, and clinical
seizure detection. Korczowski2014b (BI2014b) and Vaineau2018Brain (BI2013a)
contain visual oddball stimulation, with both target and non-target stimuli
represented as event classes using 1-s annotation intervals.
Tangermann2012 (BNCI2014-001) and Faller2012 (BNCI2015-001) provide four-class
and two-class motor-imagery tasks, respectively, with annotated intervals of
4 and 5~s. BNCI2020 contains palmar and lateral grasp events annotated with
5-s intervals, while rest is treated as background. HDOP represents LED onset,
object contact, and object lift as three event classes, each assigned a
0.2-s interval. Broderick contains continuous EEG recorded during naturalistic
speech listening, with word-level temporal annotations. TUSZ (v2.0.6)
provides clinical seizure annotations mapped to four configured categories:
focal, generalized, absence, and tonic-clonic. Duplicate seizure annotations
across channels are removed before computing the statistics.

Channel counts refer to EEG channels retained by the dataset loaders.
BNCI2020 uses 12, 32, and 58 channels in its dry-, water-, and gel-electrode
groups, respectively; the dry group includes the A2 reference channel.
HDOP retains 62 EEG channels after excluding VEOG and HEOG, while TUSZ retains
17--25 EEG channels depending on the recording montage.

For the seven non-clinical datasets, we use five-fold cross-validation (CV)
grouped by subject. One fold is held out for testing, and 10\% of the remaining
subjects are reserved for validation, rounded up to the next integer.
TUSZ follows its official train/dev/eval partition, comprising 579, 53, and
43 subjects, respectively. Input windows range from 8 to 32~s for the
non-clinical datasets and span 256~s for TUSZ.

\section{Baselines}
\label{sec:baseline_implementation}
\subsection{Baseline Implementations}

We compare NeuroDETR with six event-level baselines: a dataset-aware
random predictor, two biomedical event detectors, two temporal action
detection models, and DANCE. The comparison excludes sliding-window
classifiers because evaluating densely overlapping windows repeatedly
processes the same portions of long EEG recordings.

All methods use the same dataset-specific EEG channels, preprocessing,
event annotations, class mappings, temporal units, and subject-level
training, validation, and test partitions. Model selection is performed
exclusively on the validation set. All methods share the event-level
evaluation protocol described below.

For all dense detectors (EventNet, ActionFormer, DyFADet, and DOSED),
we apply a confidence threshold of 0.05 and class-wise non-maximum
suppression (NMS) with a temporal IoU threshold of 0.5.
After NMS, we retain at most $N_e$ predictions per window,
ranked by confidence.

\paragraph{Random.}
We construct a dataset-aware random predictor to provide a non-trivial
lower bound. For each training fold, the event-class distribution,
event-duration distribution, and event occurrence rate are estimated
using only the training annotations. The number of predicted events in
each evaluation segment is sampled according to the estimated occurrence
rate. Their classes and durations are sampled from the corresponding
training distributions, while their onset locations are sampled uniformly
within the segment. Predicted intervals are clipped to the valid temporal
range and evaluated using the same event-level protocol as the learned
models.

\paragraph{EventNet.}
EventNet denotes our implementation of the event-based modeling framework
introduced by Seeuws et al.~\citep{seeuws2024avoiding}. The original method
represents each event using its temporal center and duration. A
sequence-to-sequence backbone produces a center-confidence signal and a
duration estimate at every temporal position. During inference, local
maxima in the center-confidence signal are selected as event centers, and
the corresponding duration predictions are converted into temporal
intervals.

We extend the original binary formulation with class-specific center
and duration heads for multiclass EEG event detection. Candidate
intervals are decoded independently for each class, allowing events
of different classes to overlap. The decoded intervals undergo
the shared post-processing described above.

\paragraph{ActionFormer.}
ActionFormer~\citep{zhang2022actionformer} is a single-stage, anchor-free
temporal detector originally developed for temporal action localization
in untrimmed videos. It constructs a multiscale temporal feature pyramid,
uses local self-attention to model temporal context, and predicts an event
class together with the distances from each positive temporal location
to the left and right event boundaries.

For EEG adaptation, we replace the video feature extractor with a
one-dimensional EEG encoder. Its output is arranged as a temporal feature
sequence and processed by the original multiscale detection architecture.
The temporal strides and regression ranges of the feature pyramid are
adjusted according to the EEG sampling resolution and the
dataset-specific event-duration distribution. Predicted boundary
distances are converted into onset--offset intervals, followed by
the shared post-processing.

\paragraph{DyFADet.}
DyFADet~\citep{yang2024dyfadet} extends temporal action detection with
dynamic feature aggregation. Its dynamic encoder adjusts the aggregation
weights and temporal receptive fields according to the input features.
The associated dynamic detection head further adapts its parameters
across different feature-pyramid levels, allowing the model to handle
events with substantially different durations.

We use the same EEG encoder, input resolution, and temporal feature
interface as in the ActionFormer adaptation. The video feature sequence
is replaced by the EEG temporal representation, while the dynamic encoder,
multiscale feature pyramid, and boundary-regression mechanism are retained.
The supported regression ranges are rescaled to the event durations of
each EEG dataset. The output intervals are decoded in seconds and
processed using the shared post-processing.

\paragraph{DOSED.}
DOSED~\citep{chambon2019dosed} is a one-dimensional, anchor-based detector
originally designed for detecting multiple sleep micro-events in EEG.
The model associates a collection of default temporal events with
different positions and durations. For every default event, it jointly
predicts a class score and localization offsets. Training matches
ground-truth events to default events according to temporal overlap and
optimizes the classification and localization objectives jointly.

We adapt DOSED to the event categories of each dataset using its
default-event detection formulation. Default-event durations are
estimated using only the training annotations of each fold.
Predicted offsets are decoded into onset--offset intervals,
clipped to the input duration, and processed using the shared
post-processing.

\paragraph{DANCE.}
DANCE~\citep{levy2026dance} is a query-based model for detecting and
classifying temporal events in EEG recordings. An EEG encoder first maps
the input signal into a temporal representation. A fixed set of learnable
queries then interacts with the encoded features, with each query
predicting an event category and its temporal extent. During training,
bipartite matching establishes a one-to-one assignment between predicted
and annotated events, while unmatched predictions are assigned to the
no-event category.

In the query-sensitivity analysis, we vary only the number of decoder
queries. The capacity used to encode ground-truth events remains fixed
across the sweep, and all query settings use identical training and
evaluation annotations.

\paragraph{Unified evaluation.}
All methods ultimately output a collection of
$(t_{\mathrm{start}}, t_{\mathrm{end}}, c, s)$ tuples, where $c$ denotes
the event class and $s$ is the confidence score. Predictions are sorted
by confidence and greedily assigned to unmatched ground-truth intervals
according to temporal IoU. A matched pair contributes a true positive
only when its class is correct and its IoU satisfies the evaluation
threshold. We report event-level F1 at a temporal IoU threshold of $0.5$.


\section{Backbone Architecture}
\label{app:backbone}

We use a SimpleConv backbone adapted from~\citep{defossez2023decoding} to extract spatial and temporal features from EEG recordings. The backbone first performs spatial channel mixing, where Fourier embeddings of electrode coordinates determine learnable attention weights over the recording channels. This operation maps the input signals to a fixed set of virtual channels, incorporating electrode positions into the spatial representation. A pointwise convolution then projects the mixed signals into the feature space used by the temporal stack. 

Temporal features are extracted through ten convolutional stages with increasing dilation rates, progressively expanding the temporal receptive field. The first nine stages use 512 hidden channels with batch normalization, GELU activation, and dropout, while the final stage produces 128-dimensional features. All temporal convolutions preserve the sequence length, maintaining alignment between the extracted features and the input EEG samples for subsequent event localization and classification. Table~\ref{tab:backbone} summarizes the module configurations and input--output dimensions, where $B$, $C$, and $T$ denote the batch size, number of EEG channels, and sequence length, respectively.

\begin{table*}[!htbp]
    \centering
    \caption{Backbone architecture and hyperparameters.
    The temporal resolution is preserved throughout the backbone.}
    \label{tab:backbone}
    \small
    \renewcommand{\arraystretch}{1.2}
    \begin{tabularx}{\textwidth}{@{}lccX@{}}
        \toprule
        Module & Input shape & Output shape & Configuration \\
        \midrule
        Spatial channel mixing
        & $B\times C\times T$
        & $B\times270\times T$
        & Fourier embedding dimension: 2048;
          spatial dropout radius: 0.2 \\

        Pointwise projection
        & $B\times270\times T$
        & $B\times256\times T$
        & Conv1d, kernel size: 1 \\

        Input dropout
        & $B\times256\times T$
        & $B\times256\times T$
        & Dropout probability: 0.1 \\

        Temporal stage 1
        & $B\times256\times T$
        & $B\times512\times T$
        & Kernel size: 9; dilation: 1 \\

        Temporal stages 2--9
        & $B\times512\times T$
        & $B\times512\times T$
        & Kernel size: 9;
          dilation: $d_\ell=\lfloor2.5^{\ell-1}\rfloor$ \\

        Temporal stage 10
        & $B\times512\times T$
        & $B\times128\times T$
        & Kernel size: 9; dilation: 3814;
          no normalization or activation \\

        Transpose
        & $B\times128\times T$
        & $B\times T\times128$
        & Feature sequence for downstream modules \\
        \bottomrule
    \end{tabularx}

    \vspace{3pt}
    \begin{minipage}{\textwidth}
        \footnotesize
        All temporal convolutions use stride 1 and symmetric padding
        $4d_\ell$. Stages 1--9 each apply Conv1d, BatchNorm1d, GELU,
        dropout ($p=0.1$), and a pointwise convolution followed by
        ReLU. Stage 10 contains only the temporal convolution.
    \end{minipage}
\end{table*}

\section{Implementation Details}
\label{sec:implementation_details}
We implement NeuroDETR using PyTorch and PyTorch Lightning.
EEG signals are resampled to 128~Hz and normalized using robust
scaling. For the seven cross-validation datasets, we perform
five-fold subject-wise cross-validation. Within each fold,
10\% of the remaining subjects are reserved for validation.
TUSZ uses its predefined training, development, and evaluation
splits. The random seed is fixed to 42. We train with AdamW and a OneCycle learning-rate schedule,
using the first 10\% of training steps for learning-rate increase.
Unless otherwise specified, the peak learning rate is
$10^{-4}$, weight decay is $10^{-3}$, and the maximum gradient
norm is 1. Training runs for up to 100 epochs with an
early-stopping patience of 20 epochs.
Checkpoint selection and early stopping are based on validation
$F1_{\mathrm{event}}$ at a temporal IoU threshold of 0.5. All experiments use FP32 precision.
For the seven cross-validation datasets, we train the model on a single GPU.
TUSZ uses distributed data parallel training on eight GPUs,
with four windows per GPU and an effective batch size of 32.
\end{document}